%% file: main.tex
\documentclass[letterpaper]{article} %
\usepackage[]{aaai23}  %
\usepackage{times}  %
\usepackage{helvet}  %
\usepackage{courier}  %
\usepackage[hyphens]{url}  %
\usepackage{graphicx} %
\usepackage{natbib}  %
\usepackage{caption} %
\usepackage{newfloat}
\usepackage{listings}
\usepackage{array}
\usepackage{amsmath}
\usepackage{graphicx}
\usepackage{todonotes}
\usepackage{tikz}
\usepackage{float}
\usepackage{enumitem}
\usepackage{placeins}
\usepackage{subcaption}

\usetikzlibrary{arrows}
\usepackage{url}
\usepackage{booktabs}

\usepackage[parfill]{parskip}

\usepackage[utf8]{inputenc}
\usepackage{pgfplots}
\DeclareUnicodeCharacter{2212}{−}
\usepgfplotslibrary{groupplots,dateplot}
\usetikzlibrary{patterns,shapes.arrows}
\pgfplotsset{compat=newest}

\usepackage{adjustbox}
\usepackage{wrapfig}

\usepackage{amsmath}
\usepackage{amsthm}

\usepackage{mathtools}
\usepackage{algorithm}
\usepackage[noend]{algpseudocode}
\usepackage{bbm}

\usepackage{lipsum}  

\usepackage{rotating}

\usepackage{amsmath}
\usepackage{amssymb}
\usepackage{mathtools}
\usepackage{amsthm}
\usepackage{bm}

\usepackage{hyperref}

\usepackage{todonotes}

\title{A Coreset Learning Reality Check}
\title{A Coreset Learning Reality Check}
\author {
    Fred Lu\textsuperscript{\rm 1, 2, 3},
    Edward Raff\textsuperscript{\rm 1, 2, 3},
    James Holt \textsuperscript{\rm 3}
}
\affiliations {
    \textsuperscript{\rm 1} Booz Allen Hamilton\\
    \textsuperscript{\rm 2} University of Maryland, Baltimore County\\
    \textsuperscript{\rm 3} Laboratory for Physical Sciences\\
    Lu\_Fred@bah.com, Raff\_Edward@bah.com, holt@lps.umd.edu
}

\begin{document}

\maketitle

\begin{abstract}
Subsampling algorithms are a natural approach to reduce data size before fitting models on massive datasets. In recent years, several works have proposed methods for subsampling rows from a data matrix while maintaining relevant information for classification. While these works are supported by theory and limited experiments, to date there has not been a comprehensive evaluation of these methods. In our work, we directly compare multiple methods for logistic regression drawn from the coreset and optimal subsampling literature and discover inconsistencies in their effectiveness. In many cases, methods do not outperform simple uniform subsampling.
\end{abstract}

\section{Introduction} \label{sec:intro}

It has become common, even routine, for researchers to have massive datasets in reach for modeling and prediction. Using standard machine learning tools for datasets with over millions of observations $n$ or thousands of features $d$ is challenging when many algorithms have superlinear scaling with the data dimensions. In terms of time and resources, even training a simple linear model becomes an expensive endeavor. 

A natural workaround for excessively large data is to subsample a set of rows $m$ without losing significant information. The specific form of information to be preserved, and hence the best strategy to accomplish this, depends on the task. In the logistic regression setting, we have a data matrix $X\in \mathbb{R}^{n\times d}$, labels $y\in\{-1, 1\}^n$, and a loss function $f(x) = \log(1 + e^{-x})$, and the goal is to find the weights $\beta^\star$ which minimizes the objective function $L(\beta; X, y) = \sum_{i=1}^n f(y_i x_i^\top\beta)$. There are three natural quantities which a subsampling method may choose to approximate:
\begin{enumerate}
\item The model fit, measured as the sum of validation losses
\item The maximum likelihood estimate (MLE) $\hat\beta_{MLE}$
\item The model's performance on unseen data
\end{enumerate}
By subsampling, a practitioner will fit a logistic regression over only $m$ points instead of $n$. Each of the $m$ points is permitted an instance weight $w_i$ for the fitting procedure, which indicates the level of representation of that point in the linear model.

Recently a number of works proposed coresets for logistic regression which target quantity (1). A coreset consists of a set of $m\ll n$ weighted points whose objective function approximates that of the full dataset for any $\beta$. More precisely, if there is a subset $C$ of the data consisting of $x_{c_1}, \ldots, x_{c_m}$ and weights $w_1, \ldots, w_m$, such that for all $\beta \in \mathbb{R}^d$,
$$ \bigg|\sum_{i=1}^m w_i f(y_{c_i} x_{c_i}^\top \beta) - L(\beta; X, y) \bigg| \leq \epsilon \cdot L(\beta; X, y)$$
then the points form an $(1+\epsilon)$-coreset of $X$.

The coreset literature suggests sampling strategies, known as \textit{sensitivity sampling}, over the rows in $X$ so that with probability $1-\delta$ the resulting subsample is a $(1+\epsilon)$-coreset. The sensitivity can be viewed intuitively as the worst-case influence of a given data point on the objective over all values of $\beta$. Points which are highly influential should be more likely to be sampled into the coreset. While the sensitivity is intractable because it involves optimizing over all values of $\beta$, various strategies are used to approximate (and more specifically, upper-bound) the sensitivity in order to sample data. These include $k$-means clustering \cite{huggins2016coresets}, leverage scores \cite{munteanu2018coresets}, and row norms \cite{tolochinsky2018coresets}. Using theoretical frameworks based on \cite{feldman2011unified,braverman2016new}, probabilistic bounds are derived to relate a given subsample size $m$ to $\epsilon$ and $\delta$ (for example, the minimum $m$ needed to achieve some $\epsilon$).

Because the primary quantity of interest has been approximating the objective function, attention has not been given to how the minimizer $\hat\beta_C$ of the coreset objective relates to $\hat\beta_{MLE}$. In particular, the standard $(1+\epsilon)$-coreset needs to hold for all $\beta$, whereas we may only be interested in the approximation performance of the coreset in a neighborhood of $\hat\beta_{MLE}$. Recently the statistics literature on optimal subsampling has investigated this area by focusing on target (2). \citet{wang2018optimal} proposes a subsampling procedure with conditions where $\hat\beta_C$ converges to $\hat\beta_{MLE}$. While the asymptotic performance of such procedures has been studied \cite{wang2018optimal,ting2018optimal}, finite sample bounds like those in the coreset literature have not yet been developed.

Finally, while some of the above works have empirically examined coreset performance on (3), this has not been a main focus. Since large-scale logistic regression models are often trained in order to get predictive power over unseen data, it is crucial that prediction quality does not greatly deteriorate when subsampling is applied.

Based on these frameworks, numerous sampling strategies have been presented in the literature with theoretical support. While each respective method was supported by experimental results, a comprehensive evaluation has been missing. Some limitations of previous studies are:
\begin{itemize}
    \item Each method is benchmarked with at most one or two competitors on a small selection of real-world datasets. The datasets vary greatly in size depending on the study, from thousands of rows to millions.
    \item The evaluation metric has not consistent across these works, with some assessing objective function error, some measuring the error of estimating $\hat\beta$ itself, etc.
    \item Subsample size ranges: Some works examine coresets with only hundreds of samples even on large datasets, while others do not consider very small coreset sizes. In our work we find that some methods are more effective in certain size regimes.
\end{itemize}

In our experiment design we account for these limitations by benchmarking most known subsampling methods for logistic regression over a large variety of realistic datasets. We are the first to present a thorough empirical comparison of these approaches, over a range of important metrics. In the process we discover that most existing algorithms show inconsistencies in performance across datasets, with many of them not outperforming the uniform sampling baseline. Furthermore, we find that one of the earliest coreset methods still performs among the best, despite being ignored in all recent evaluations. We additionally show that our results are robust across regularization settings.

In \autoref{sec:related_work} we give context on the importance of reproducibility research. We then detail the coreset and optimal subsampling algorithms in \autoref{sec:methods} and our evaluation procedure in \autoref{sec:evaluation}. Our results are shown and discussed in \autoref{sec:results}. Finally we conclude in \autoref{sec:conclusion}.

\section{Related Work} \label{sec:related_work}

The primary focus of our study is on the matter of reproducible research. Considerable work has been done in replicating research as described by the original manuscripts \cite{Raff2019_quantify_repro} and in developing repeatable infrastructure so that a manuscript's code can be easily re-run \cite{Claerbout1992,Forde2018,Zaharia2018AcceleratingTM,Forde2018ReproducingML}. While such research is valuable for understanding the nature of replication and its incentives \cite{Raff2020c,Raff2022,Raff2022a}, it also repeats any methodological errors from the original manuscript where conclusions could change under different evaluation \cite{pmlr-v97-bouthillier19a}. 

This methodological concern has grown in recent years, with many sub-fields in machine learning discovering different results when a more thorough evaluation is performed. \citet{Musgrave2020} showed that metric learning was not improving with new triplet mining and losses, but rather more general neural architectures, data augmentation, and optimization strategies. \citet{mlsys2020_73} found inconsistencies in neural architecture search and built a new benchmarking framework to standardize. \citet{Dacrema2019} and \citet{10.1145/3383313.3412489} also found that in recommender systems many results could be replicated as described in the original works, but when a simple baseline was properly tuned the advantage of proposed techniques disappeared. This is similar to our own work, except our baseline is naive uniform random sampling. \citet{Eggensperger2021} similarly showed hyper-parameter optimization conclusions could change in a thorough benchmark, though most methods did improve upon naive baselines. Our work follows in performing a through evaluation of coreset methods, where we find that most prior seminal methods often fail to improve upon the naive baseline of uniform random sampling. 

A sub-concern of methodological evaluation is the method by which conclusions about one method being ``better'' than another are drawn. Early work in the machine learning field proposed a number of alternative schemes on how to balance valid statistical testing with the computational limits of performing the necessary experiments ~\cite{Kubat1997,Bradley1997,Dietterich:1998:AST:303222.303237,Alpaydin:1999:CTC:339993.339999}, but multiple later results~\cite{Demsar:2006:SCC:1248547.1248548,JMLR:v17:benavoli16a} showed that using a non-parametric test was a highly effective means of drawing a valid conclusion, with computational limits mostly unnecessary given advancements in computing speed. This is not true in all cases. In particular deep learning can be highly compute intensive and needs alternative strategies \cite{Bouthillier2021}, but our work is focused on logistic regression coresets and so we can use the non-parametric test for robust conclusions. While other approaches to robust conclusions have been proposed recently we consider them beyond our intended scope ~\cite{Soboroff:2018:MRE:3269206.3271719,Berrar2016}.

\section{Subsampling Methods} \label{sec:methods}

\begin{table*}[t]
    \centering
    \adjustbox{max width=\textwidth}{%
    \begin{tabular}{lccl} \toprule
        Coreset     & Weights      & Coreset size       & Details \\ \midrule
        $k$-means   & $s_i = \frac{n}{1+\sum_{j=1}^k |G_j^{(-i)}| e^{-R \lVert \bar{G}_{j}^{(-i)} - Z_i\rVert} }$       & $O\big(\frac{\bar{s}}{\epsilon^2} (d+1)\log \bar{s} + \log\frac{1}{\delta}   \big)$        & $G_j$ are $Z_i$ assigned to cluster $j$, $Z_i=X_i y_i$ \\
        \texttt{Leverage} & $s_i = \lVert Q_i\rVert_2 + \frac{1}{n}$, where $Z=QR$ & $\tilde{O}\big(\frac{\mu_y(X) \sqrt{n}  d^{3/2} }{\epsilon^2} \big)$ & $Z_i=-X_i y_i$, $X$ is $\mu$-complex\\
        \texttt{Monotonic}   &  $s_i=\frac{132\sqrt{k} \lVert p_i \rVert + 2}{i}$    & $O\big( \frac{t}{\epsilon^2}(d\log t + \log\frac{1}{\delta} ) \big)$   & $p_i=\mathrm{sorted}(x_i)$, $k=\frac{1}{2\lambda}$, $t=\sum s_i$ \\
        \texttt{Lewis} & $s_i=\mathrm{Lewis}(X)$  &   $\tilde{O}\big( \frac{\mu_y(X)^2 d }{\epsilon^2}   \big)$  & $\mu$-complex, \texttt{Lewis} in \cite{cohen2015lp} \\
        \texttt{OSMAC} (vc) & $s_i = |y_i - p_i(\hat\beta)| \cdot \lVert x_i\rVert$ & -- & $p_i$ estimated probs. using pilot $\hat\beta$\\
        \texttt{OSMAC} (mse) & $s_i = |y_i - p_i(\hat\beta)| \cdot \lVert M_X^{-1} x_i\rVert$ & -- & $M_X=\frac{1}{n} \sum p_i(\hat\beta)(1-p_i(\hat\beta)) x_i x_i^\top$ \\
        uniform & $s_i = 1$ & $\tilde{O}\big( \frac{n^{1-\kappa} d}{\epsilon^2}  \big)$ &  $\lVert x_i\rVert_2 \leq 1$, regularization $\lambda \propto n^\kappa$ \\ \bottomrule
    \end{tabular}
    }
    \caption{Computational expressions (corresponding to line 1 of Algorithm \ref{algo:procedure}) and theoretical size of a $(1+\epsilon)$-coreset for each coreset method evaluated. The notation $\tilde{O}$ suppresses logarithmic factors. This is used in some methods like \texttt{Leverage} where the $\log 1/\delta$ is re-expressed as a multiplicative logarithmic constant, or the expression is otherwise very lengthy. Throughout $X\in \mathbb{R}^{n\times d}$, $y \in \mathbb{R}^n$,  $\lambda$ is the regularization parameter, and $\delta$ is the failure probability.}
    \label{tab:method_details}
\end{table*}

We identified and implemented several subsampling methods for logistic regression from the literature. All methods follow the same general procedure, as described in Algorithm \ref{algo:procedure}. The final weights are designed so that the sum of losses over the subsample is on the scale of the loss over all $n$ points, which enables the coreset objective to approximate the full data objective. The subsample is then used to fit a weighted logistic regression model, i.e. $\hat\beta_C$ minimizing $\sum_{j=1}^m w_i f(y_{c_j} x_{c_j}^\top \beta)$. This subsampled estimate is finally compared with the full data MLE. We next briefly describe the methods being evaluated along with implementation details that we found were necessary in order to reproduce the original results. We refer the reader to the original publications for the nuanced details of each method. This six prior methods, plus our naive baseline are detailed below. Furthermore, mathematical expressions for the coreset weights and theoretical sizes are shown in \autoref{tab:method_details}.

\begin{algorithm}[h]
\centering
\caption{Coreset sampling procedure}
\label{algo:procedure}
\begin{algorithmic}[1]
\Require Datasets $X\in \mathbb{R}^{n\times d}$, $y\in \mathbb{R}^n$, desired subsample size $m$, method to compute importance weights $\mathrm{CoresetMethod}$
\State Compute $\{s_i\}_{i=1}^n \gets \mathrm{CoresetMethod}(X, y)$
\For{$i$ in $1:n$}
    \State $p_i \gets s_i / \sum_{i=1}^n s_i$
\EndFor
\State Sample $m$ rows from $X, y$ using probabilities $\{p_i\}$ to get $X_c, y_c$
\For{$j$ in $1:m$}
    \State $w_j \gets 1 / (p_{c_j} \cdot n)$
\EndFor
\State \Return coreset $(X_c, y_c, w) = \{x_{c_j}, y_{c_j}, w_j\}_{j=1}^m$
\end{algorithmic}
\end{algorithm}

\textbf{$k$-means coreset} \cite{huggins2016coresets}. This was one of the first coresets proposed for logistic regression. The theoretical analysis is based on the framework developed in \cite{braverman2016new, feldman2011unified}, where the key step is to upper bound the sensitivity score of each data point. If the upper bound is too loose, then a larger $m$ is needed to achieve the same $\epsilon$ coreset. This work used $k$-means clustering on a uniform subsample of the data to compute the upper bound importance weights based on distance from cluster centers. The method has two hyperparameters, the number of clusters $k$ and the radius $R$ of the ball containing $\beta$. We found that the results were robust to $k$ and chose $k=6$. While the theory requires $\beta$ to be bounded for the results to hold, in practice we found $R=1$ usually performed the best even if $\max_d |\hat\beta_{MLE}|$ exceeded 1.

The only other work to evaluate the $k$-means coreset is \cite{munteanu2018coresets}, where $R$ was set to be unbounded. However, as  $R\to\infty$ the importance weights converge to the uniform distribution, so their results are not representative of the effectiveness of the $k$-means method. No other works have benchmarked this method, so we felt it was important to give it a fair chance.

\textbf{Leverage coreset} \cite{munteanu2018coresets}. The $\ell_2$-leverage scores have been used to approximate $\lVert X \theta\rVert_2$ with $\lVert X' \theta \rVert_2$ for a row-wise reduced matrix $X'$ up to $(1+\epsilon)$ relative error \cite{cohen2015lp}. In \cite{munteanu2018coresets}, the root leverage scores are shown to be an effective upper bound on the sensitivities. The scores are computed by taking the row-wise L2-norm of an orthonormal basis of $X$. Such a basis was obtained with the QR-decomposition. A further step bins the sampling probabilities into multiples of 2 to reduce the complexity of the procedure. 

While an approximate QR-decomposition algorithm was used in the paper, it was faster for us to use the optimized \texttt{numpy.linalg.qr} routine for an exact solution. Another important detail for the method to give good results was to make sure any columns of 1s used to include the intercept were removed prior to computing $Q$.

\textbf{Monotonic coreset} \cite{tolochinsky2018coresets}. Unlike the previous coreset works, this work develops bounds on the sensitivity of logistic regression when regularization is used. The sampling weights are computed to be proportional to the lengths of the data points, scaled by the relative index of each point in sorted order. The weights are further scaled by $\sqrt{k}$ where $k$ is the regularization parameter (which in our version is equivalent to $\frac{1}{2\lambda}$). Thus their framework requires the use of regularization.

\textbf{Lewis coreset} \cite{mai2021coresets}. An $\ell_1$ analog of the leverage scores, the \texttt{Lewis} scores approximate the quantity $\lVert X\theta\rVert_1$ up to $(1+\epsilon)$ relative error. Using their similarity to hinge-like loss functions including the logistic loss, this work derives coreset results for such functions. While intuitively similar to the root leverage coreset, they are the first to show a linear rather than polynomial dependence on the feature dimension $d$ via their proof technique. In practice, computing the \texttt{Lewis} weights requires iteratively computing the leverage scores $t$ times using the algorithm of \cite{cohen2015lp}, thus giving an approximate runtime of $t$ times that of the \texttt{Leverage} coreset. As reported in their paper, we found that the method fully converged for $t=20$ while being acceptable for $t=5$, so we used $t=5$ to avoid excessive run times.

\textbf{OSMAC} \cite{wang2018optimal}. This work shows that under relatively mild conditions, the MLE $\hat\beta$ under a subsampling distribution converges to the full data MLE as the coreset size increases. The \texttt{osmac\_{mse}} is proposed as such a subsampling method. Furthermore the distribution of $\hat\theta - \hat\theta_{MLE}$ is shown to be asymptotically normal, with the resulting covariance matrix being minimized by the \texttt{OSMAC\_{MSE}} weights. The scores are computed by taking into account the residual $(y - \hat y)$, normalized by a covariance matrix weighted by the expected binomial variance at each point. This has a similar form to the leverage while using label and model fit information, and can also be interpreted as the influence function as in \cite{ting2018optimal}. The authors additionally propose a second method \texttt{OSMAC\_{VC}} which replaces the covariance matrix with the identity for computational efficiency.

Both methods require the actual MLE $\theta$ to estimate the residuals and weighted Hessian (in the first case) accurately, so an initial pilot estimate using a uniform subsample is required. In our experiments we used a subsample of $m/2$ for the pilot estimate. We found that results were unreliable at small sizes. Checking with the source code, we identified that the authors weighted the pilot sample to have evenly balanced classes. Implementing this ourselves led to more stable pilot estimates.

\textbf{Influence subsampling} \cite{ting2018optimal}. The influence function uses a second-order approximation of the loss to estimate the effect of infinitessimally upweighting each point in the dataset, as motivated in \cite{hampel1974influence,koh2017understanding}. For the logistic regression case, the form of the influence is equivalent to the score used in \texttt{OSMAC\_{MSE}}. Thus we do not show this method separately. Like the previous, this work also proves the asymptotic optimality of influence sampling, albeit with a different framework.

\textbf{Uniform coreset}. It was shown in \cite{munteanu2018coresets} that in unconstrained logistic regression there exist general datasets for which no sublinear coreset exists \cite{munteanu2018coresets}. In order to bound the sensitivity, they rely on the notion of $\mu$-complexity, which measures for a dataset, over all $\beta$, the worst-case ratio of sum-log-odds between incorrectly and correctly classified points. The smaller this quantity is, the more effective the \texttt{Leverage} coreset is. \cite{curtin2019coresets} pointed out that when sufficient regularization is used, the sensitivity of logistic regression is controlled over all datasets. Therefore, uniform subsampling is an effective coreset for all datasets and is in fact nearly optimal under sufficient regularization.

In our experiments we use very weak regularization and will show that even in nearly unconstrained settings that the uniform coreset is competitive with other methods.

\textbf{Combining coresets}. Coreset methods can be composed, e.g. by recursively applying the subsampling procedure. For example, \cite{munteanu2018coresets} give additional theoretical bounds for a recursive coreset. Due to the added complexity when stacking results, we consider these methods to be outside the scope of our evaluation. Thus in all methods we limit the coreset generation to a single round of subsampling.

\section{Evaluation procedure} \label{sec:evaluation}

\begin{table}[ht]
    \centering
    \begin{tabular}{lcccc}
    \toprule
        Dataset & $n$ & $d$ & $d$ (num.) & \% pos  \\ \midrule
        chemreact & 24059 & 100 & 100 & 3.00 \\
        census & 30932 & 100 & 6 & 24.1 \\
        bank & 39128 & 51 & 8 & 11.3 \\
        webspam & 126185 & 127 & 127 & 60.7 \\
        kddcup & 469319 & 41 & 35 & 80.3 \\
        covtype & 551961 & 54 & 10 & 51.2 \\
        bitcoin &  2770862 & 24 & 6 & 1.42 \\
        SUSY & 4500000 & 18 & 18 & 45.7 \\
        \bottomrule
    \end{tabular}
    \caption{Characteristics of datasets used in the study, with rows $n$, features $d$, numeric features (i.e. not categorical columns which are one-hot encoded), and percent positive class. Additional details on dataset preprocessing and sources are in the Appendix.}
    \label{tab:data}
\end{table}

Our aim is to unify previous experimental methods into a single comprehensive procedure. In this procedure we aim to resolve the limitations from previous evaluations:

\textbf{Datasets}. While previous work used at most 3 datasets, we evaluate on 8 datasets which include previously used ones as well as new ones. The sizes range from 24000 to nearly 5 million (\autoref{tab:data}).

\textbf{Evaluation metric}. In accordance with the three aims of subsampling as described in the Introduction, we assess the following metrics:
\begin{itemize}
    \item Relative negative log-likelihood (NLL) error to assess the model fit: $$|L(\hat\beta_C; X, y) - L(\hat\beta_{MLE}; X, y)| / L(\hat\beta_{MLE}; X, y)$$
    \item The mean squared error (MSE) of the MLE:
    $$\lVert \hat\beta_C - \hat\beta_{MLE}\rVert_2^2$$
    \item Relative ROC of the subsampled model on validation data: $ROC(\hat\beta_C, X, y) / ROC(\hat\beta_{MLE}, X, y)$
\end{itemize}

\textbf{Subsample sizes}. We consider 25 subsample sizes on a logarithmically spaced grid from 200 to 100000. (For datasets with fewer rows we limit the upper end of the evaluation range.) We replicate each procedure 50 times and report the medians and inter-quartile intervals. This is a much broader range of evaluation than in previous work.

\textbf{Regularization}. Most coreset methods directly handle regularization by adding the penalty to the loss function, giving for example $f(x; \beta)=\log(1 + e^{-x}) + \lambda \lVert \beta \rVert_2^2$. Some previous works, including \cite{munteanu2018coresets, wang2018optimal} only evaluate their methods on unconstrained logistic regression, while certain methods require the use of regularization \cite{tolochinsky2018coresets}.  Furthermore, recent theoretical work has shown that the uniform subsampling method produces an effective coreset in the presence of regularization \cite{curtin2019coresets}. In our main experiments, we use weak L2 regularization at $\lambda=10^{-5}$ as some of the datasets produce unstable results without any regularization. Coefficient estimates generally did not further change when further weakening $\lambda$. We conduct additional analysis at varying $\lambda$ values and our main conclusions did not change.

Our $\lambda$ is defined with respect to the objective $\frac{1}{2n} \sum f(y_ix_i^\top \beta) + \lambda \lVert \beta \rVert_2^2$. Because of the normalizing $n$, when we fit a subsampled logistic regression, we normalize the modeling weights $w_i$ to maintain the same level of regularization with respect to the data loss term. For loss parameterizations without the normalizing $n$, the original $w_i$ should be used.

\begin{figure*}
\centering
\begin{subfigure}{0.99\textwidth}
\centering
\hspace*{\fill}  Relative NLL error $(\downarrow)$ \hfill Coefficient MSE $(\downarrow)$ \hfill Relative ROC $(\uparrow)$ \hspace*{\fill}
\end{subfigure}
\begin{subfigure}{0.99\textwidth}
\begin{turn}{90} 
\phantom{BufferBuffer}Webb spam 
\end{turn}
    \centering
    \hspace*{\fill}
    \includegraphics[width=0.29\textwidth, trim=0cm 0cm 0cm 0cm]{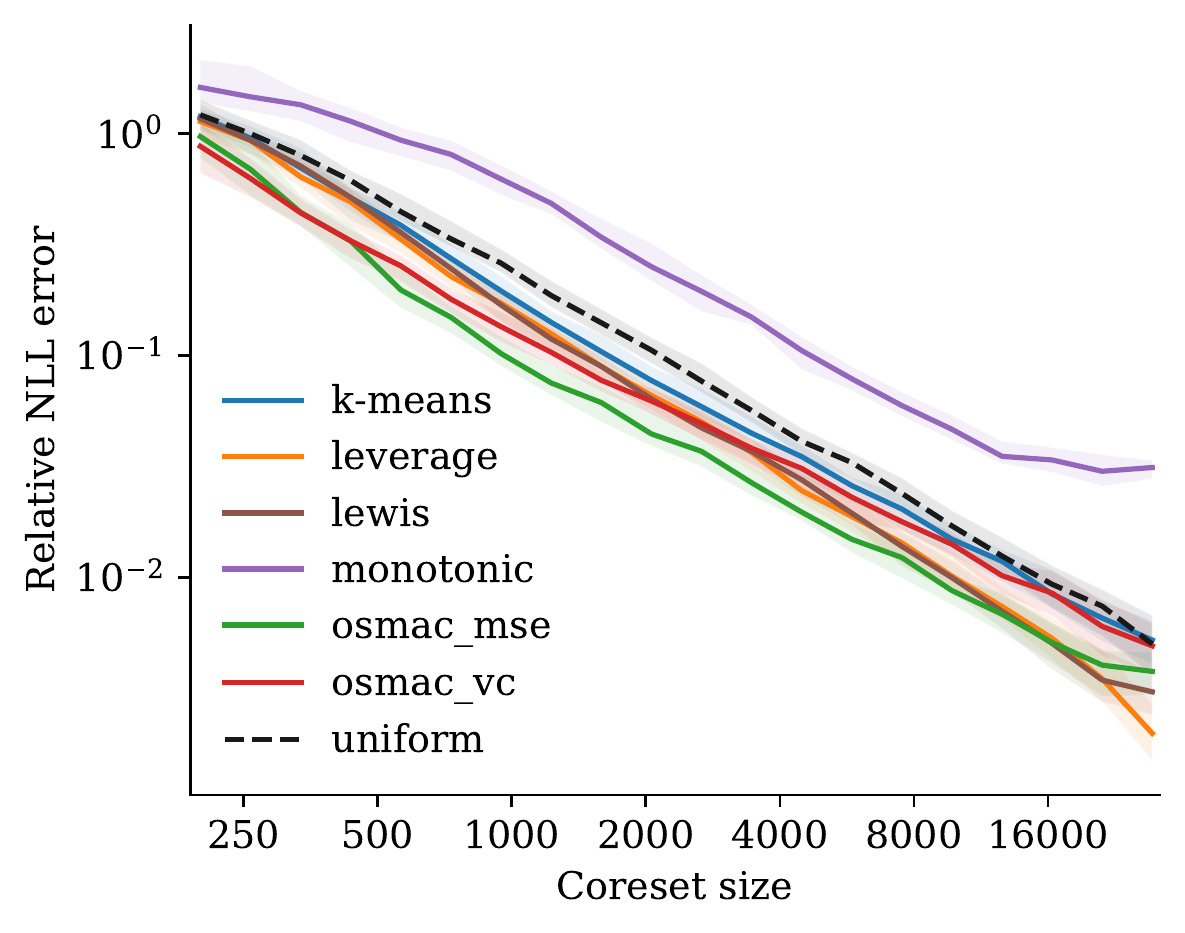}
    \hfill
    \includegraphics[width=0.29\textwidth, trim=0cm 0cm 0cm 0cm]{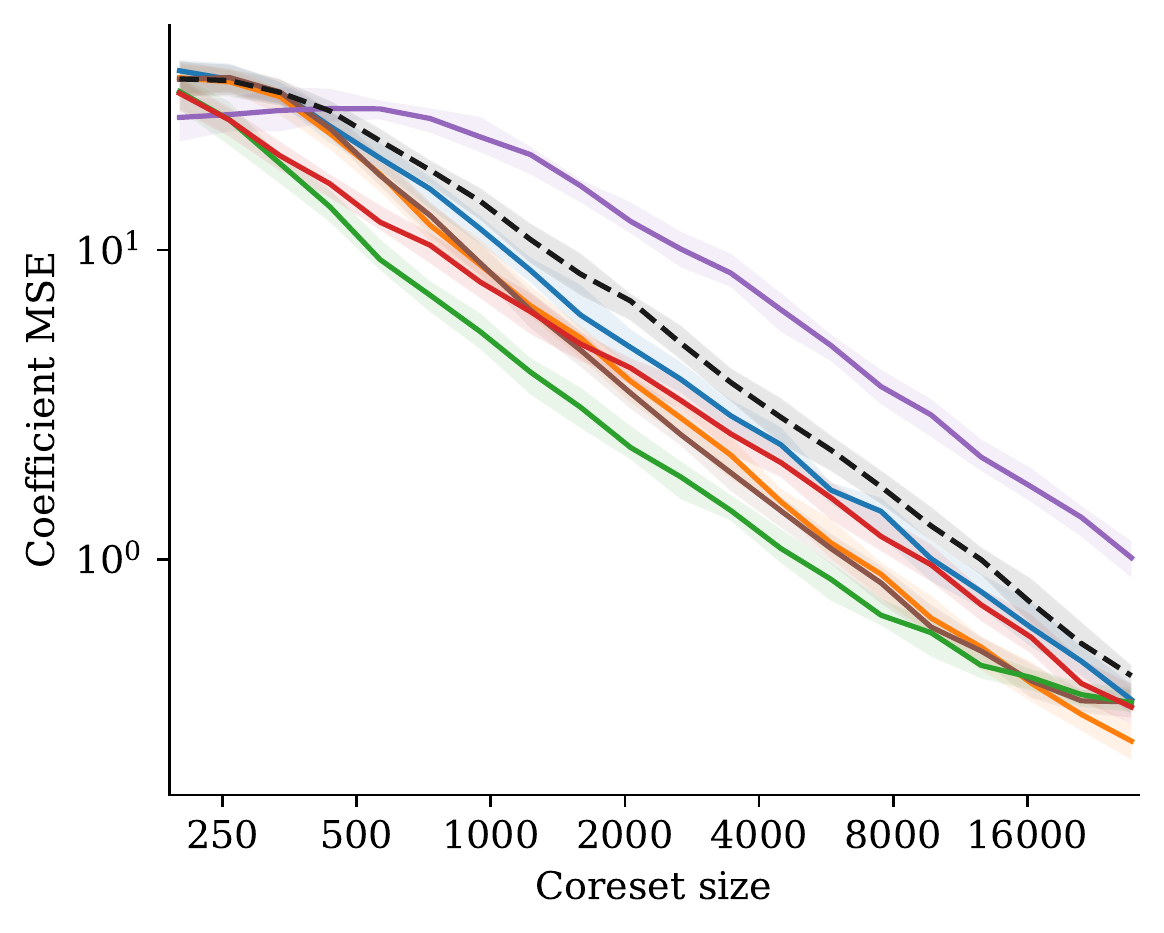}
    \hfill
    \includegraphics[width=0.29\textwidth, trim=0cm 0cm 0cm 0cm]{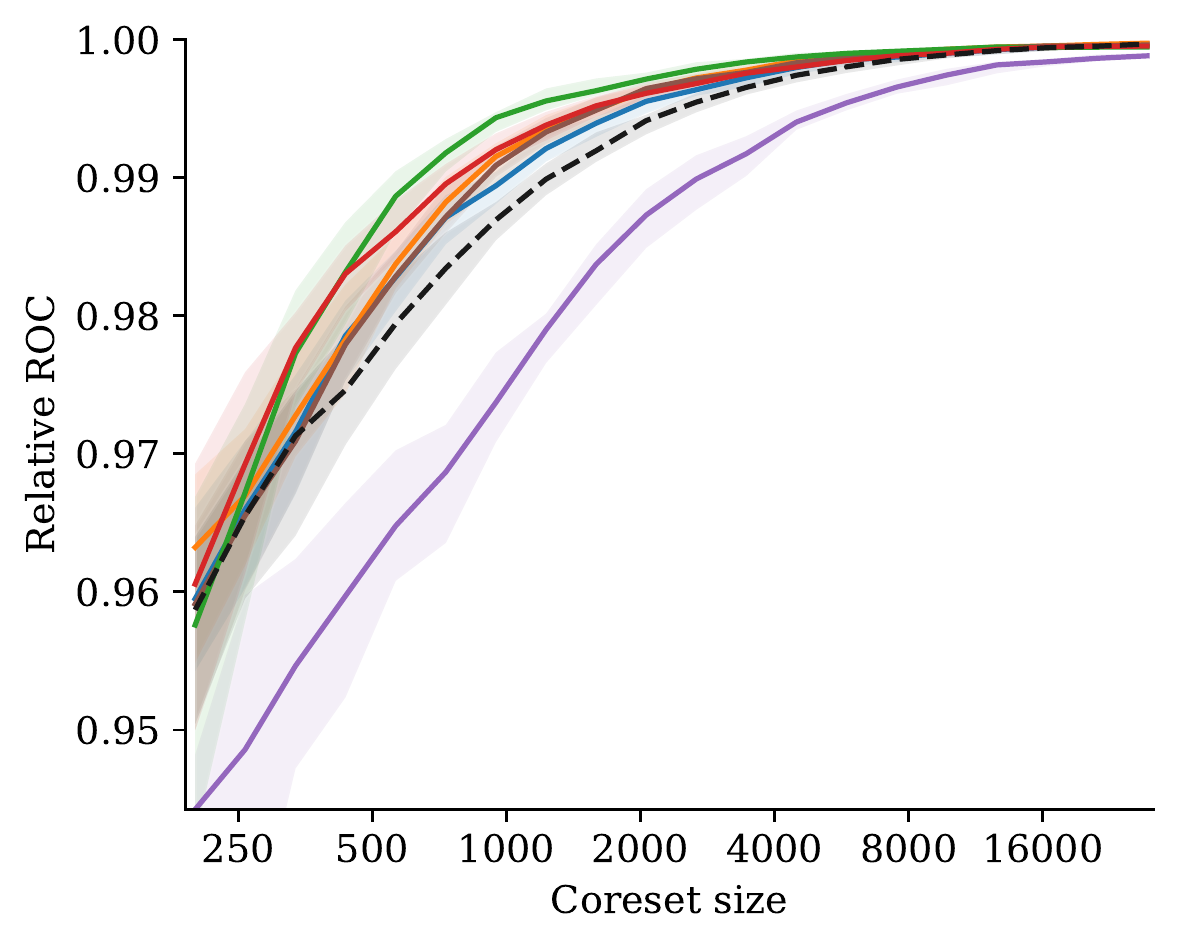}
    \hspace*{\fill}
\end{subfigure}

\begin{subfigure}{0.99\textwidth}
\begin{turn}{90} 
\phantom{BufferBuffer}KDD cup
\end{turn}
    \centering
    \hspace*{\fill}
    \includegraphics[width=0.29\textwidth, trim=0cm 0cm 0cm 0cm]{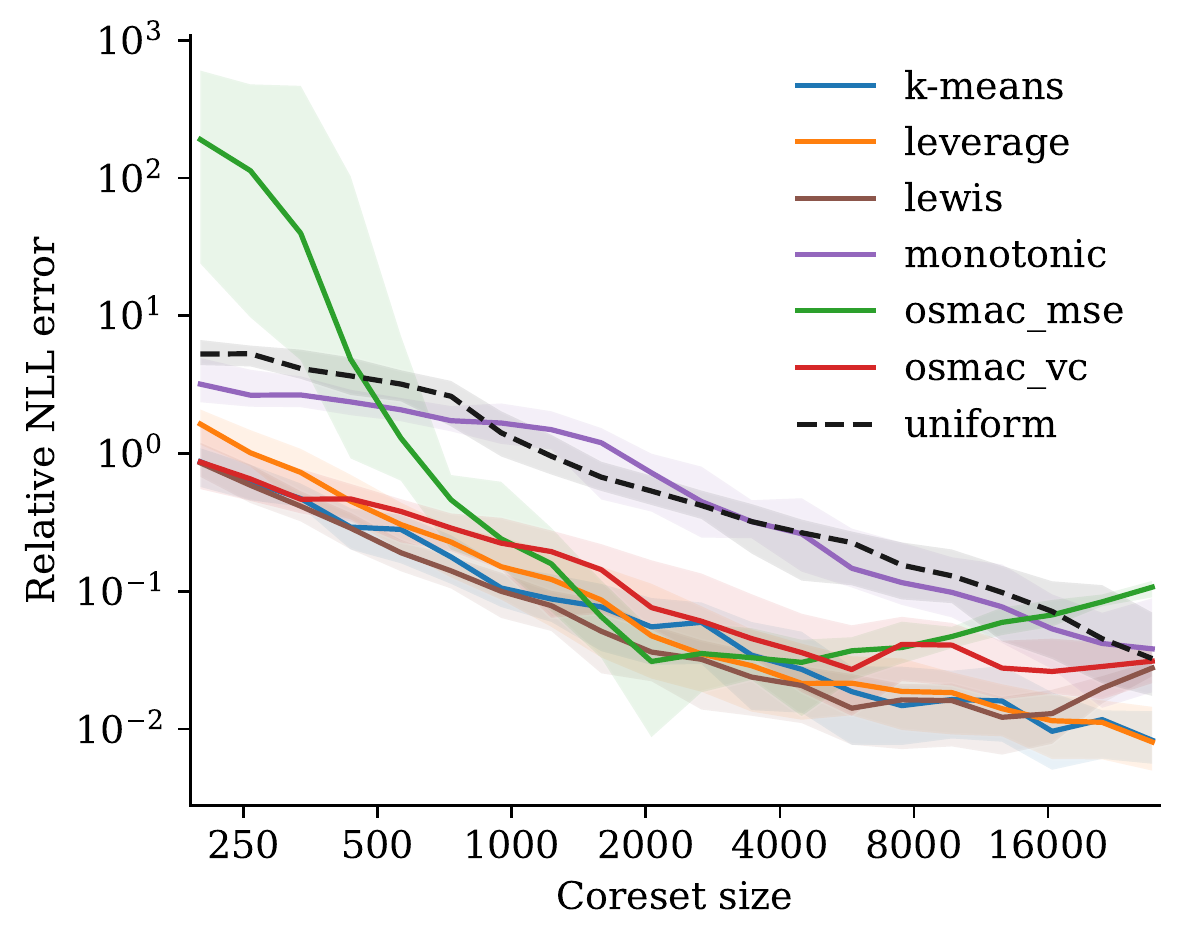}
    \hfill
    \includegraphics[width=0.29\textwidth, trim=0cm 0cm 0cm 0cm]{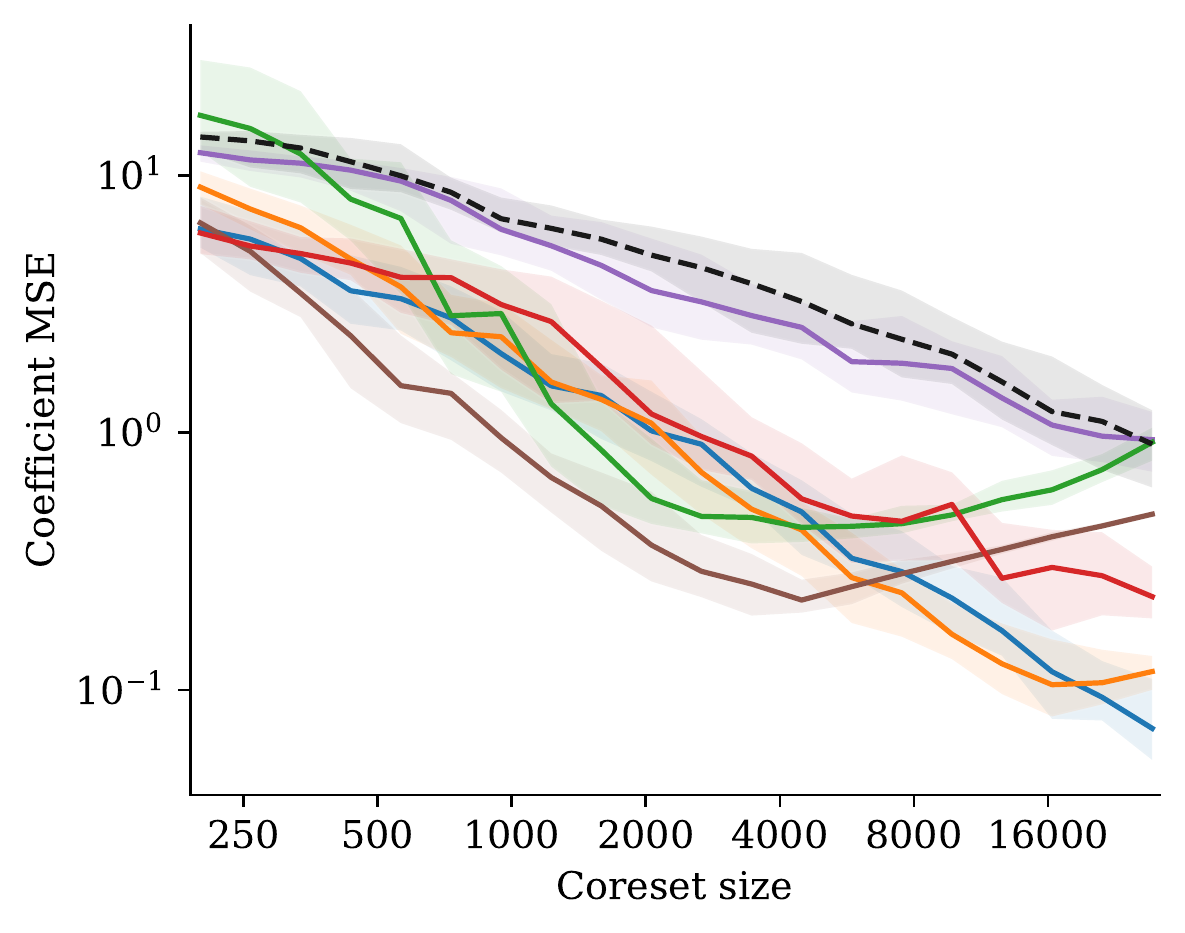}
    \hfill
    \includegraphics[width=0.29\textwidth, trim=0cm 0cm 0cm 0cm]{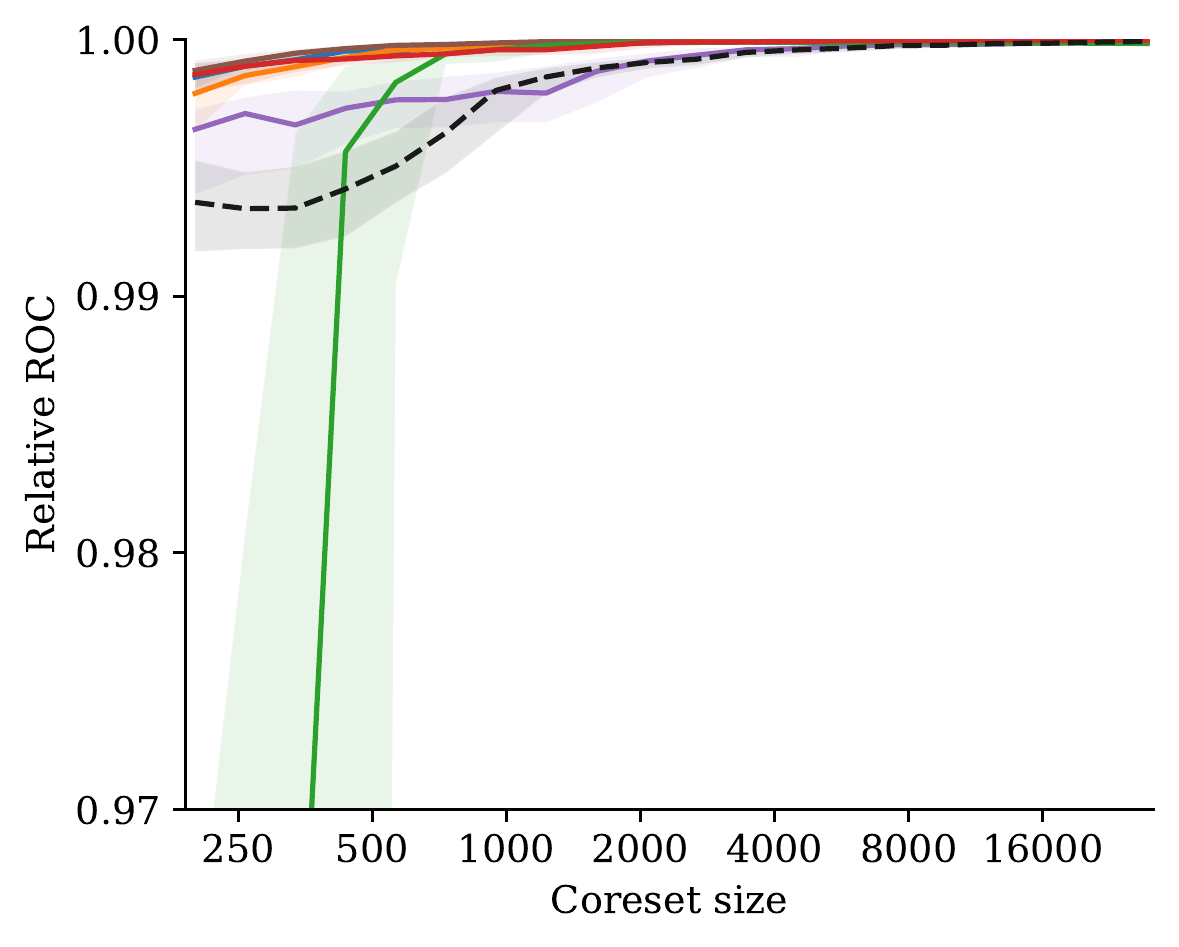}
    \hspace*{\fill}
\end{subfigure}

\begin{subfigure}{0.99\textwidth}
\begin{turn}{90} 
\phantom{BufferBuffer}covertype
\end{turn}
    \centering
    \hspace*{\fill}
    \includegraphics[width=0.29\textwidth, trim=0cm 0cm 0cm 0cm]{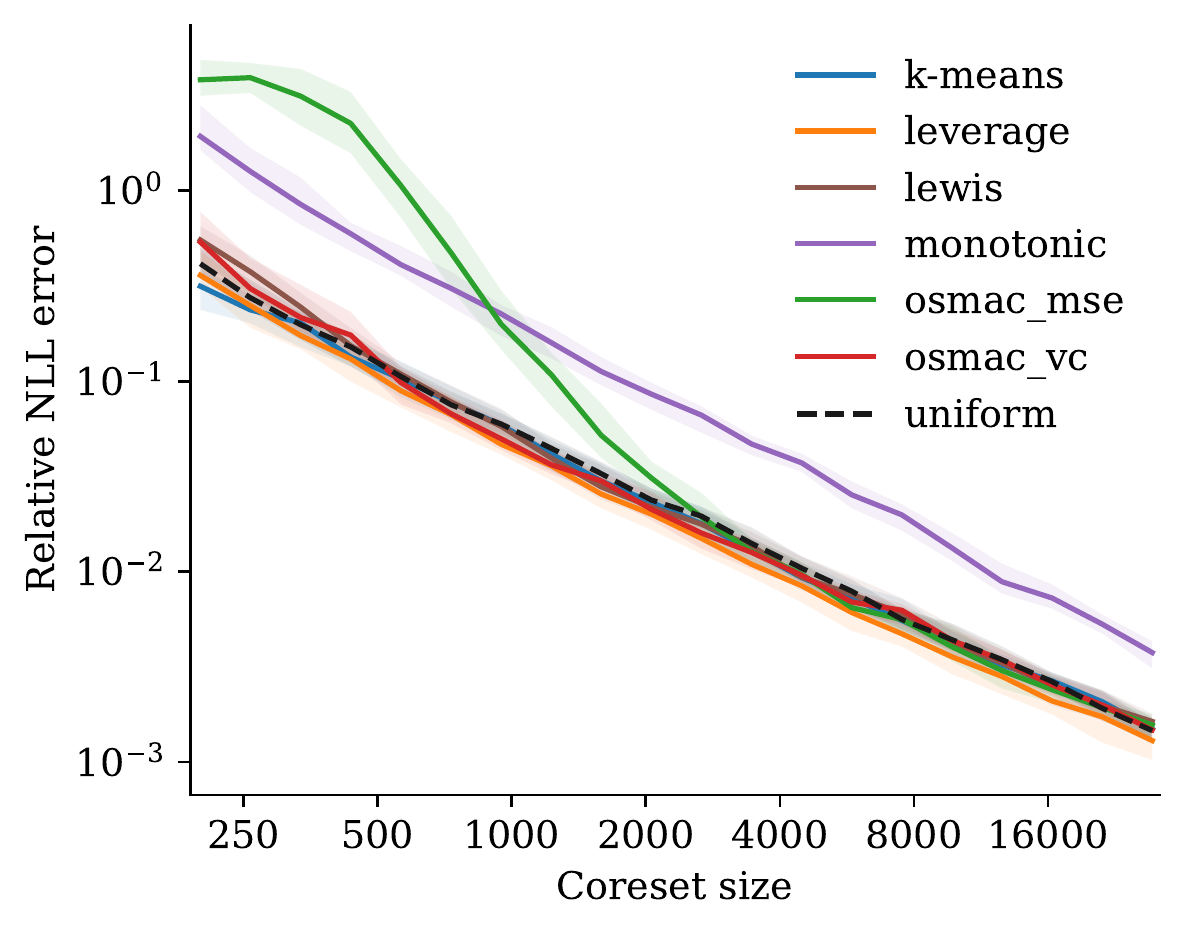}
    \hfill
    \includegraphics[width=0.29\textwidth, trim=0cm 0cm 0cm 0cm]{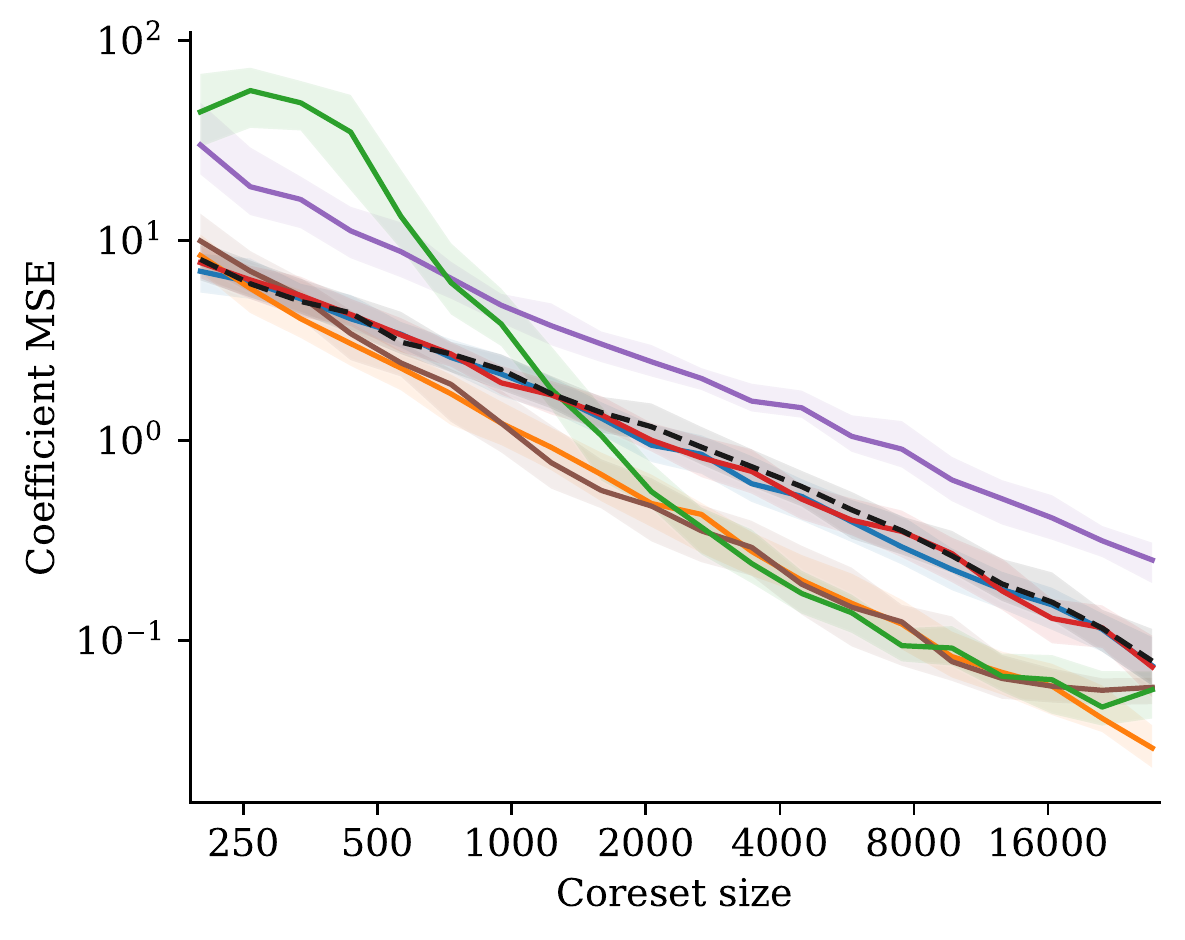}
    \hfill
    \includegraphics[width=0.29\textwidth, trim=0cm 0cm 0cm 0cm]{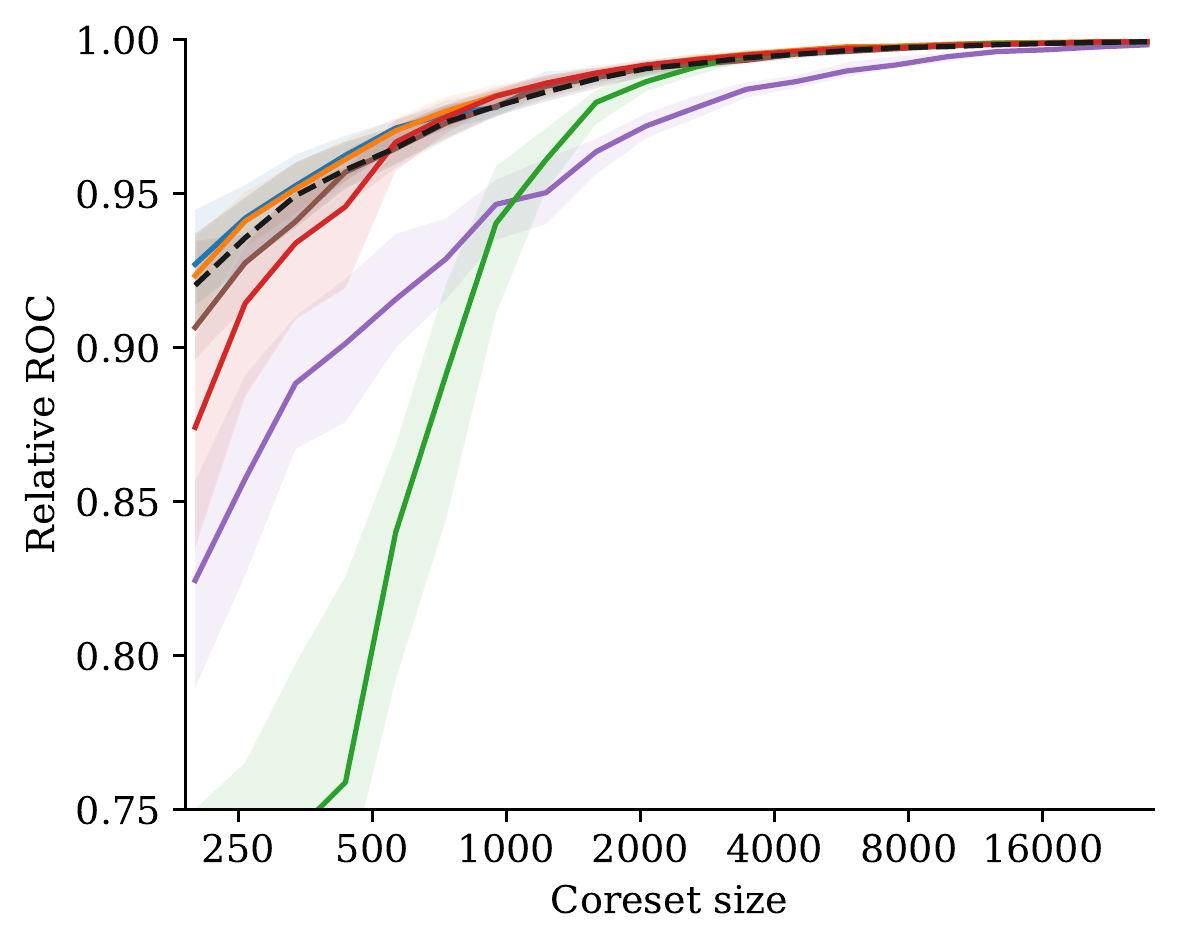}
    \hspace*{\fill}
\end{subfigure}

\begin{subfigure}{0.99\textwidth}
\begin{turn}{90} 
\phantom{BufferBuffer}bitcoin 
\end{turn}
    \centering
    \hspace*{\fill}
    \includegraphics[width=0.29\textwidth, trim=0cm 0cm 0cm 0cm]{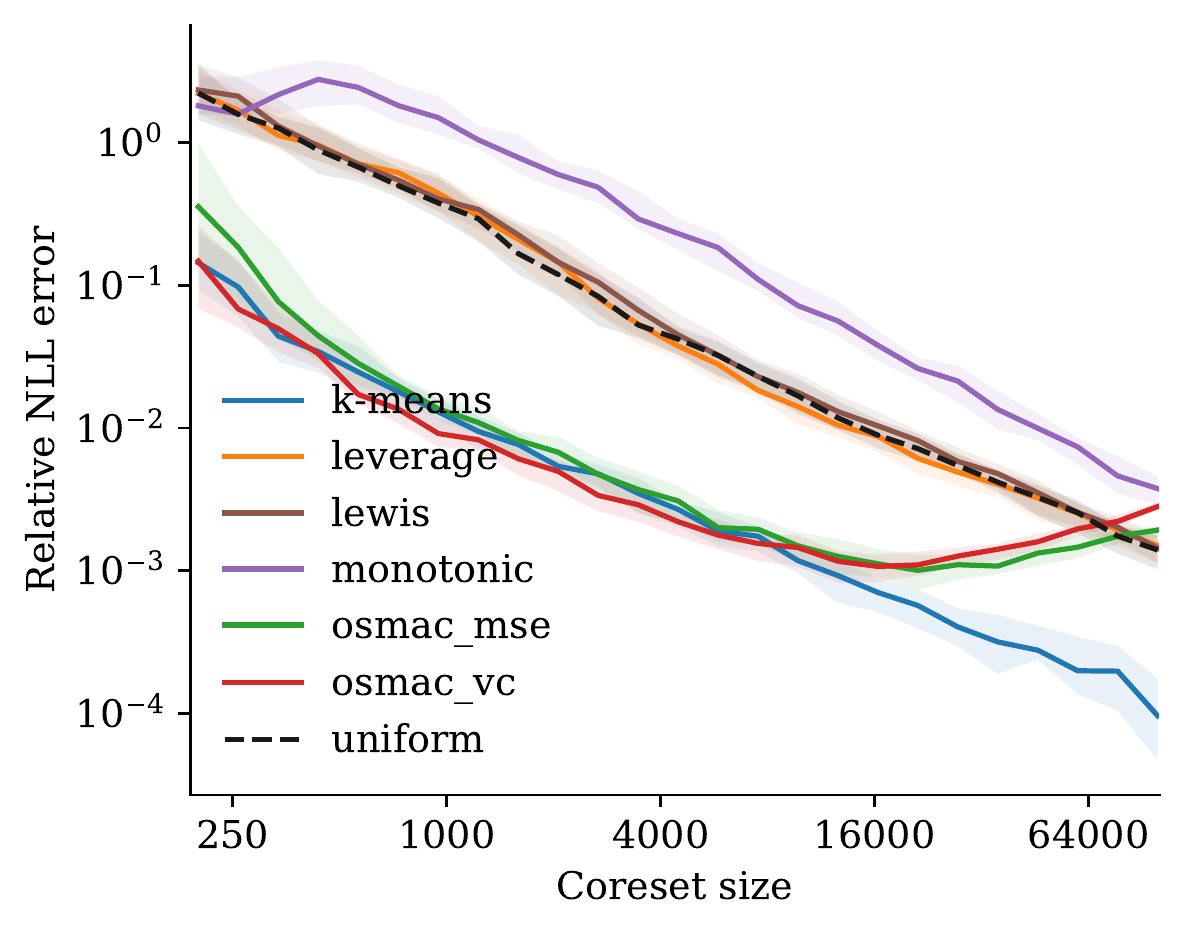}
    \hfill
    \includegraphics[width=0.29\textwidth, trim=0cm 0cm 0cm 0cm]{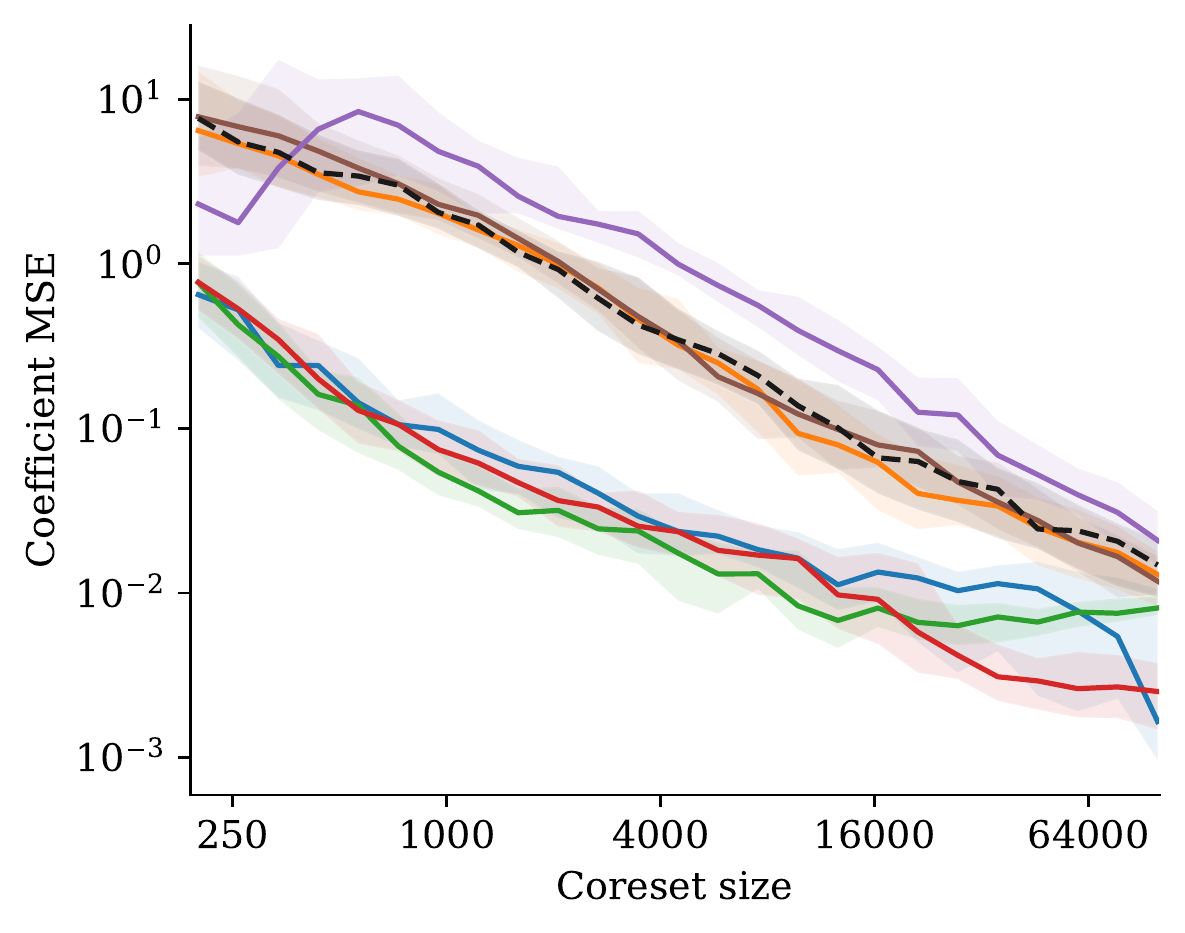}
    \hfill
    \includegraphics[width=0.29\textwidth, trim=0cm 0cm 0cm 0cm]{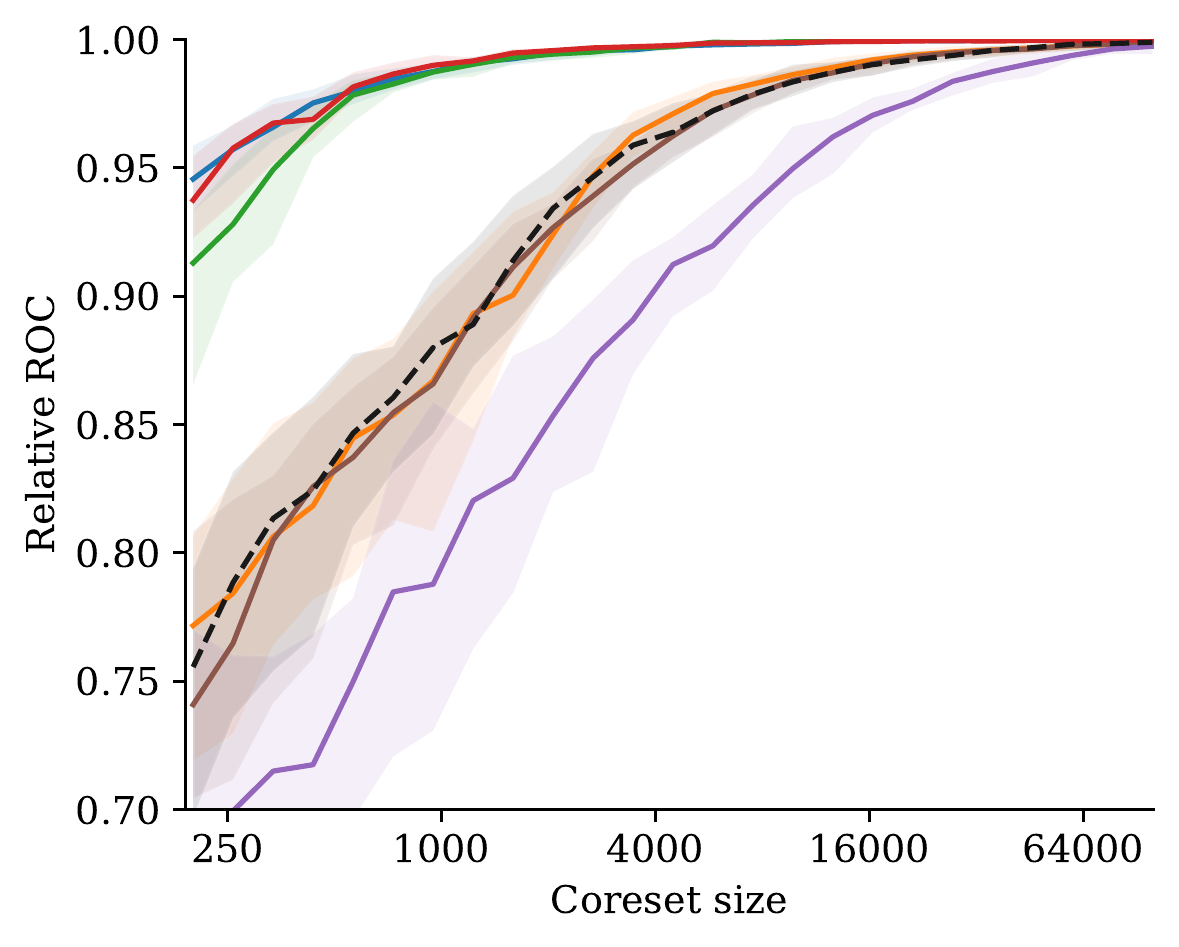}
    \hspace*{\fill}
\end{subfigure}

\begin{subfigure}{0.99\textwidth}
\begin{turn}{90} 
\phantom{BufferBuffer}SUSY
\end{turn}
    \centering
    \hspace*{\fill}
    \includegraphics[width=0.29\textwidth, trim=0cm 0cm 0cm 0cm]{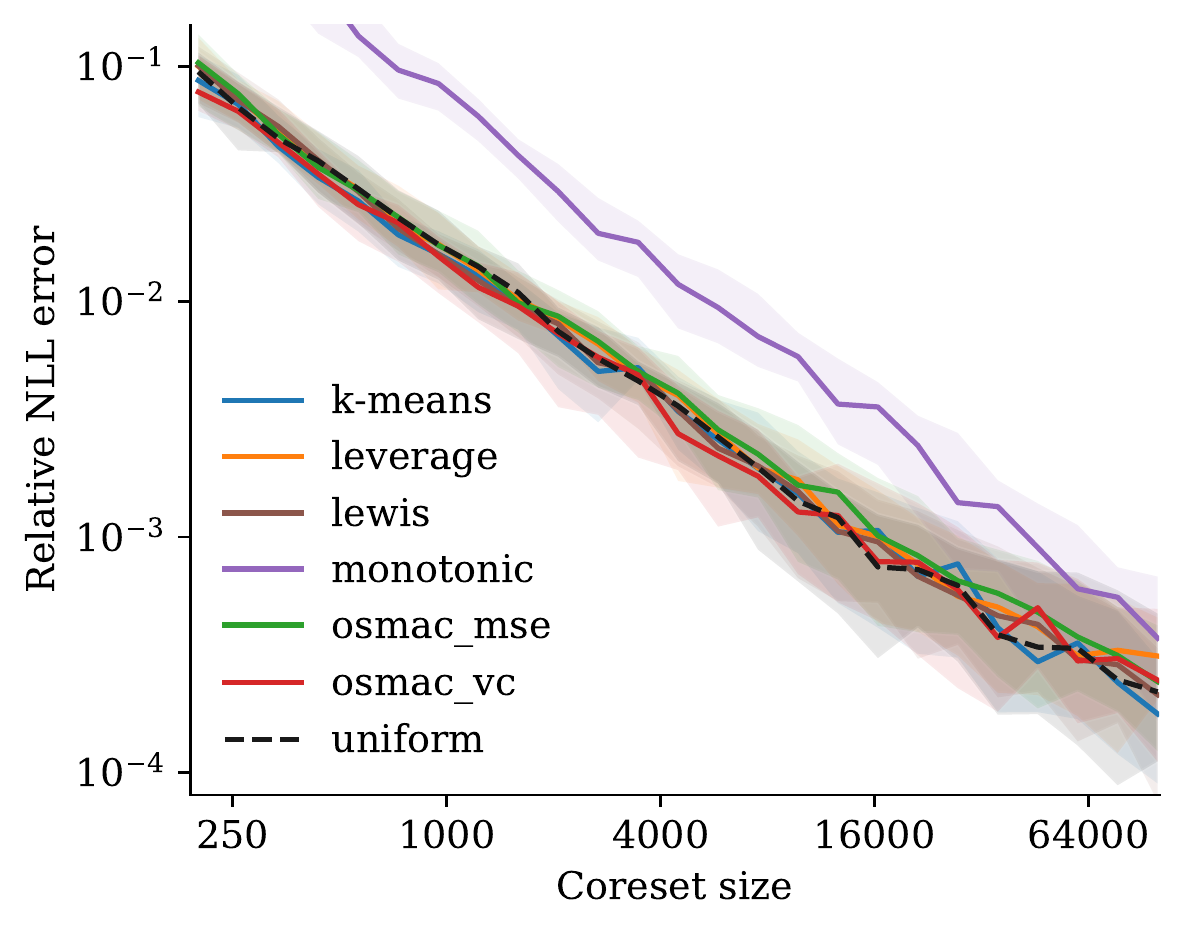}
    \hfill
    \includegraphics[width=0.29\textwidth, trim=0cm 0cm 0cm 0cm]{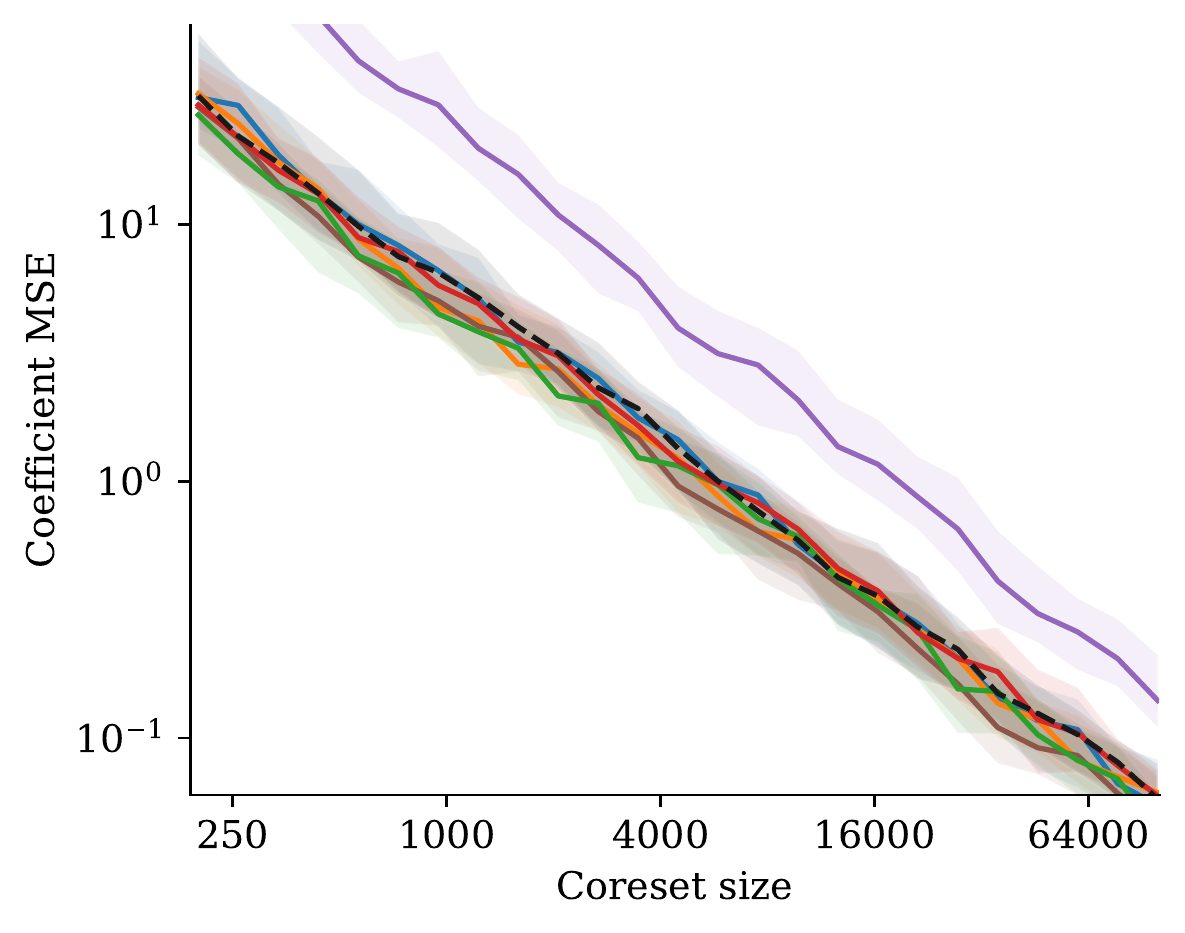}
    \hfill
    \includegraphics[width=0.29\textwidth, trim=0cm 0cm 0cm 0cm]{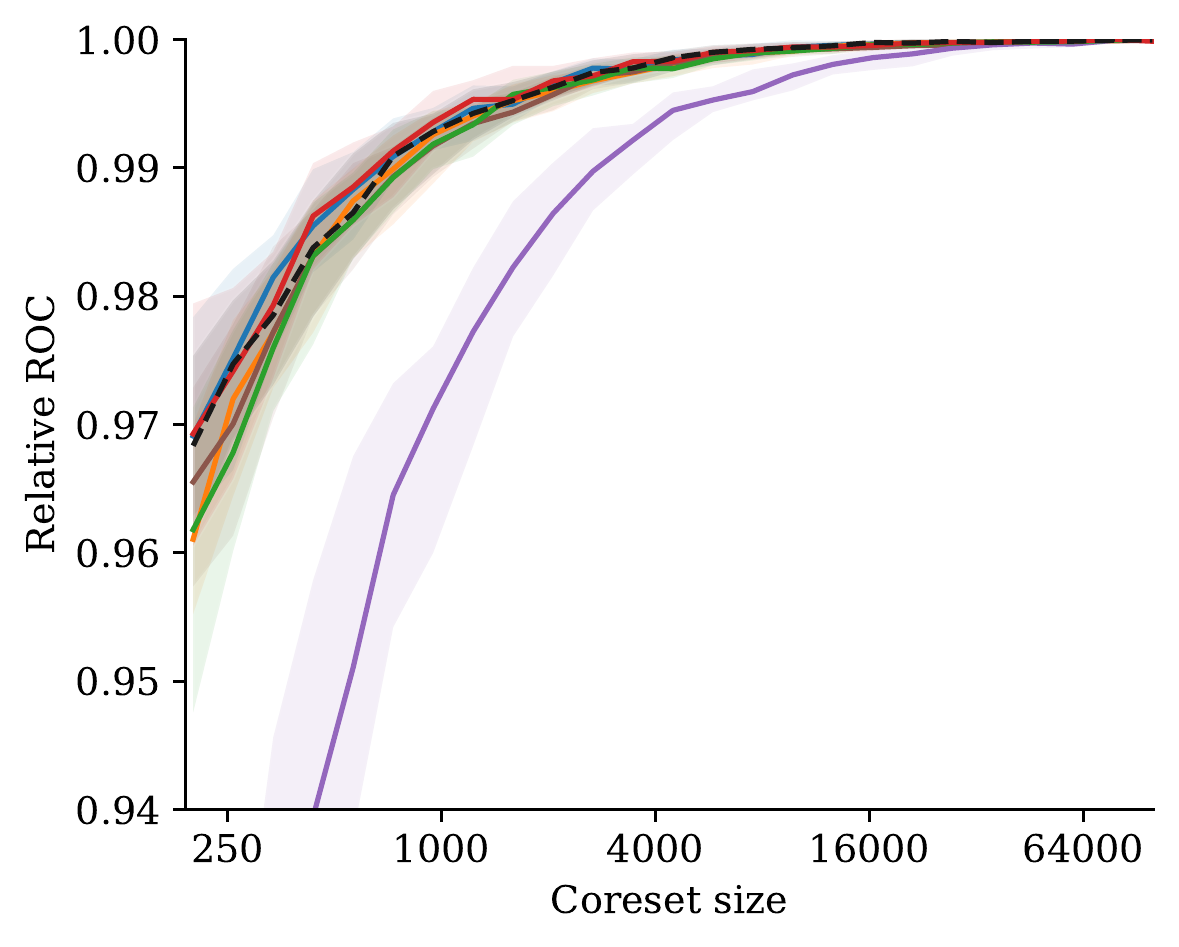}
    \hspace*{\fill}
\end{subfigure}
\caption{All subsampling methods are compared over a range of 25 subsample sizes. Each row represents one dataset and each column represents a different evaluation metric. Each point represents a median of 50 replications, with shaded areas representing interquartile range. The five largest datasets are shown here, with the remaining deferred to \autoref{fig:main_more}.}
\label{fig:main}
\end{figure*}

\section{Results} \label{sec:results}

Our primary comparison is shown in \autoref{fig:main}, with remaining datasets in \autoref{fig:main_more}. The relative performance of subsampling algorithms vary by dataset but are generally consistent across metrics. This shows that accurately estimating $\hat\beta_{MLE}$ with $\hat\beta_C$ leads to effective logistic regression subsampling regardless of the metric of interest. 

\begin{figure*}[h]
\centering
\begin{subfigure}{0.99\textwidth}
\centering
\hspace*{\fill}  $\lambda=10^{-3}$ \hfill $\lambda=10^{-1}$ \hfill $\lambda=10$ \hspace*{\fill}
\end{subfigure}
\begin{subfigure}{0.99\textwidth}
\begin{turn}{90} 
\phantom{BufferBuffer}KDD cup
\end{turn}
    \centering
    \hspace*{\fill}
    \includegraphics[width=0.29\textwidth, trim=0cm 0cm 0cm 0cm]{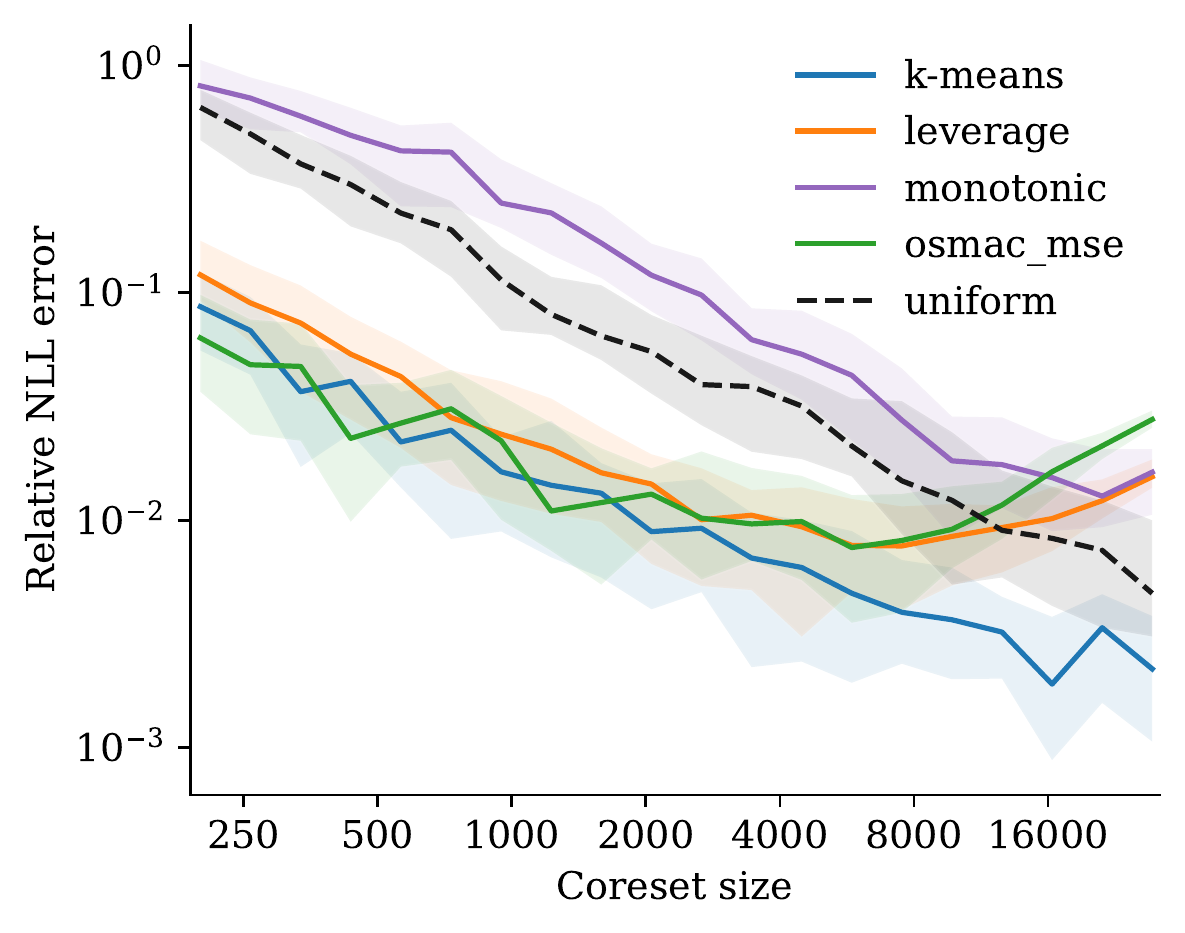}
    \hfill
    \includegraphics[width=0.29\textwidth, trim=0cm 0cm 0cm 0cm]{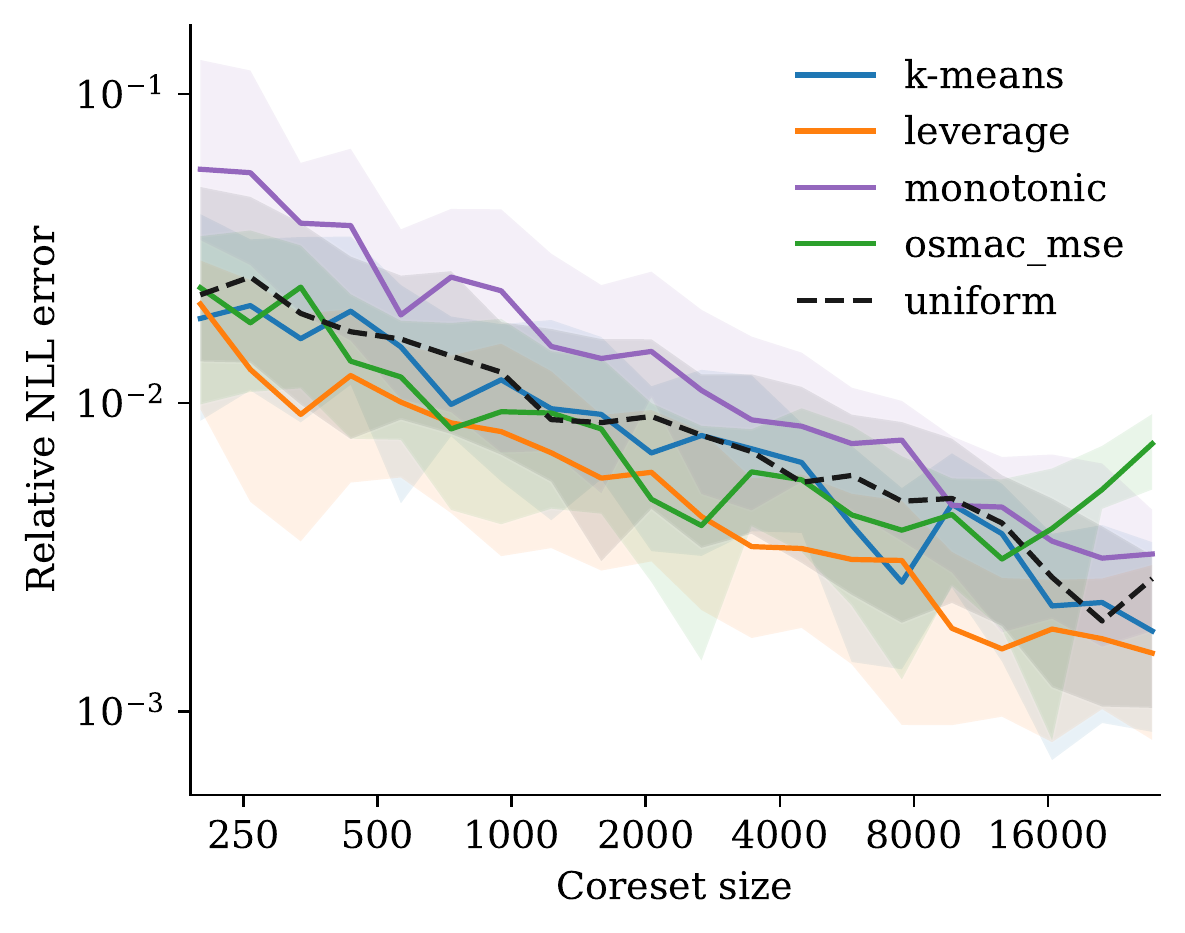}
    \hfill
    \includegraphics[width=0.29\textwidth, trim=0cm 0cm 0cm 0cm]{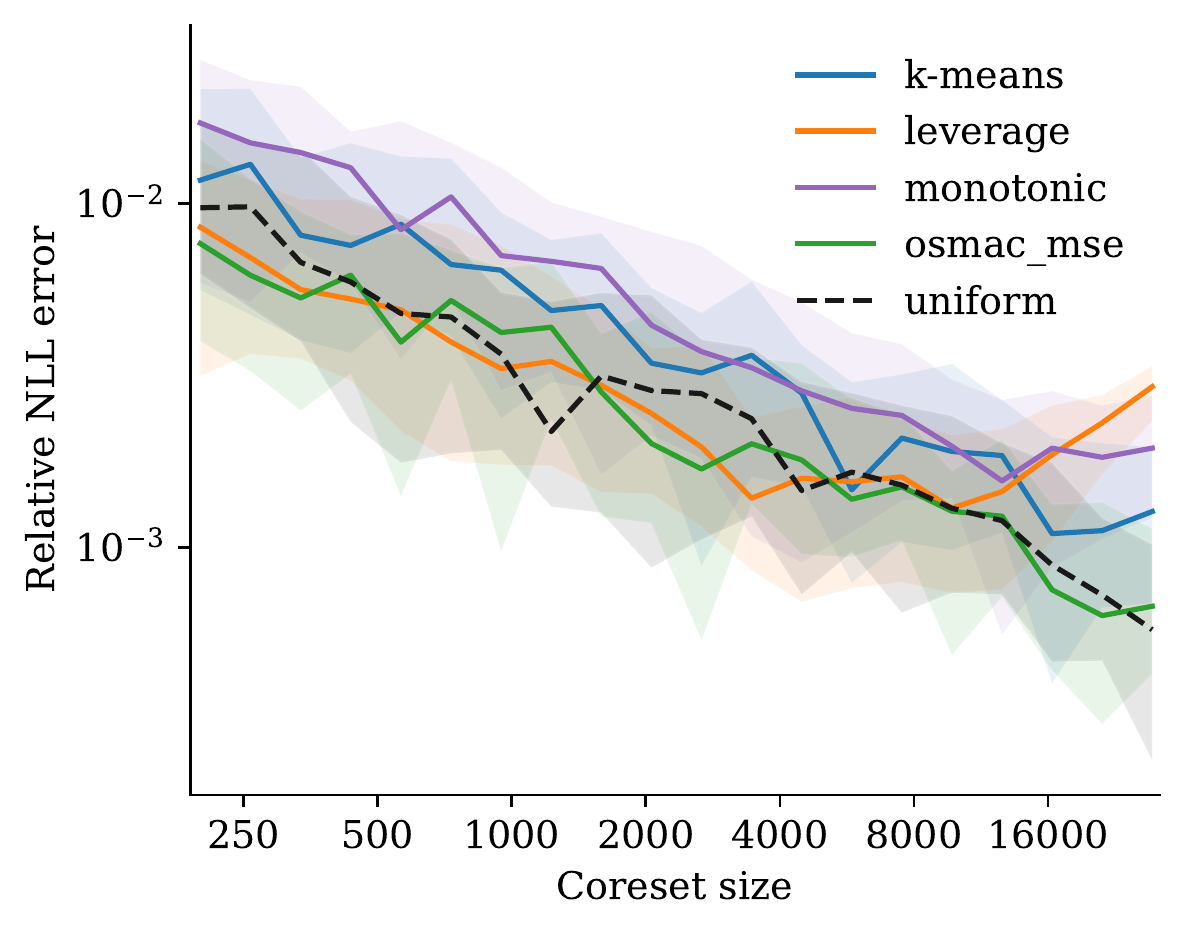}
    \hspace*{\fill}
\end{subfigure}

\caption{On the KDD Cup 99 dataset, increasing L2 regularization noticeably improves the efficiency of uniform subsampling relative to other methods. At high regularization, no advantage is observed from any other method. This shows concordance with the theoretical results in \cite{curtin2019coresets}. Similar effects can be seen on other datasets in \autoref{fig:l2_more}.}
\label{fig:l2}
\end{figure*}

Methods such as \texttt{Leverage}, \texttt{Lewis}, and \texttt{OSMAC\_{MSE}} are very effective on KDD cup relative to the \textit{uniform} baseline, but not so much on others. In fact, the uniform baseline frequently outperforms other coreset methods across datasets. In addition, while previous works which had stopped showing the \textit{$k$-means} coreset due to perceived lack of performance, our results find that despite being the earliest method, it turned out to be one of the most competitive when properly tuned. The \texttt{Monotonic} method was formulated to work with regularization, so we may expect it to perform weaker. However, increasing the L2 penalty still does not improve it over the uniform baseline (\autoref{fig:l2}). The \texttt{OSMAC} methods show large variability, with \texttt{OSMAC\_{MSE}} giving strong performances on Webb spam and bitcoin but worse performances on other datasets, especially at small subsampled sizes. This may be due to challenges in obtaining a reliable inverse Hessian matrix with small sample sizes.

We follow recommendations by \cite{Demsar:2006:SCC:1248547.1248548,JMLR:v17:benavoli16a} and use a non-parametric test to determine if there is a significant difference between all coreset methods. We compute the non-parametric Dunn test using the JASP software. This software performs a multiple test correction per metric, but not across metrics. Since we have three metrics, this means the proper threshold for significance is $\alpha \leq 0.05/3 \approx 0.0167$. We first look at the results in \autoref{tbl:uniform}, where each method is compared against the uniform sampling baseline. This result shows that when accounting for the number of tests performed, only the \texttt{OSMAC} methods beat the uniform baseline and only when measuring by MSE of $\beta$. We remind the reader this is a somewhat Pyrrhic victory, as Figures \ref{fig:main} and \ref{fig:main_more} show that both \texttt{OSMAC} variants perform worse than uniform sampling in different regimes of the coreset size.

\begin{table}[!h]
\centering
\caption{Comparisons with uniform baseline, with multiple test corrected $p$-value in the three
right most columns. We can see that none of the coreset methods are significantly different than a uniform random sampling baseline by the NLL or ROC metrics. Only \texttt{OSMAC} outperforms uniform sampling for the MSE metric. }
\label{tbl:uniform}
\begin{tabular}{@{}lccc@{}}
\toprule
\multicolumn{1}{c}{Comparison} & NLL   & $\beta$ MSE    & ROC   \\ \midrule
\texttt{k-means} - uniform          & 0.123 & 0.08           & 0.318 \\
\texttt{Leverage} - uniform         & 1.000 & 0.527          & 1.000 \\
\texttt{Lewis} - uniform            & 1.000     & 0.249          & 1.000     \\
\texttt{Monotonic} - uniform        & 0.055 & 0.29           & 0.23  \\
\texttt{OSMAC\_MSE} - uniform       & 0.758 & \textbf{0.004} & 1.000     \\
\texttt{OSMAC\_VC} - uniform        & 0.057 & \textbf{0.006} & 0.075 \\ \bottomrule
\end{tabular}
\end{table}

\begin{figure}[!h]
\begin{subfigure}{\columnwidth}
    \centering
    \includegraphics[width=\columnwidth, trim=0cm 0cm 0cm 0cm]{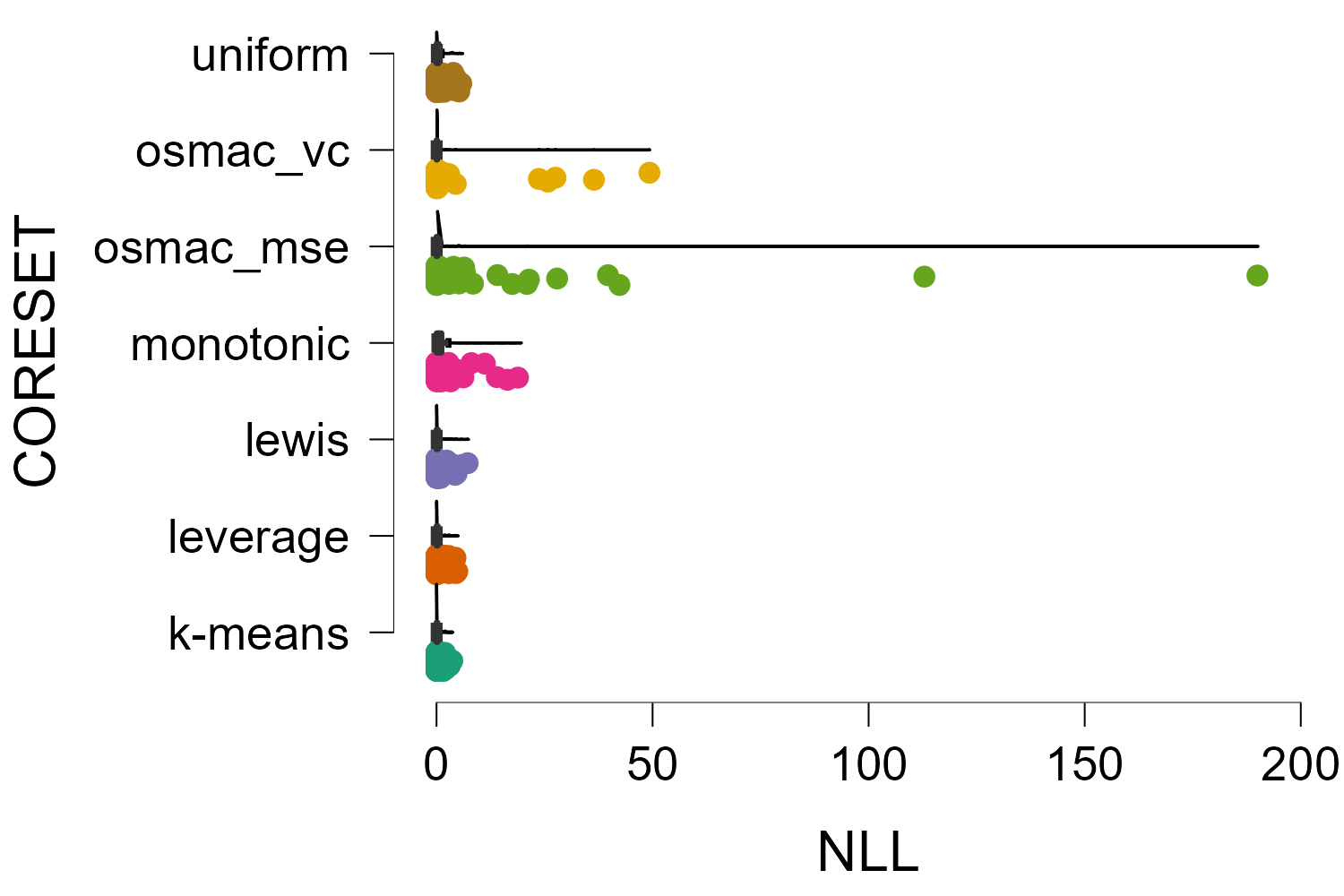}
    \includegraphics[width=\columnwidth, trim=0cm 0cm 0cm 0cm]{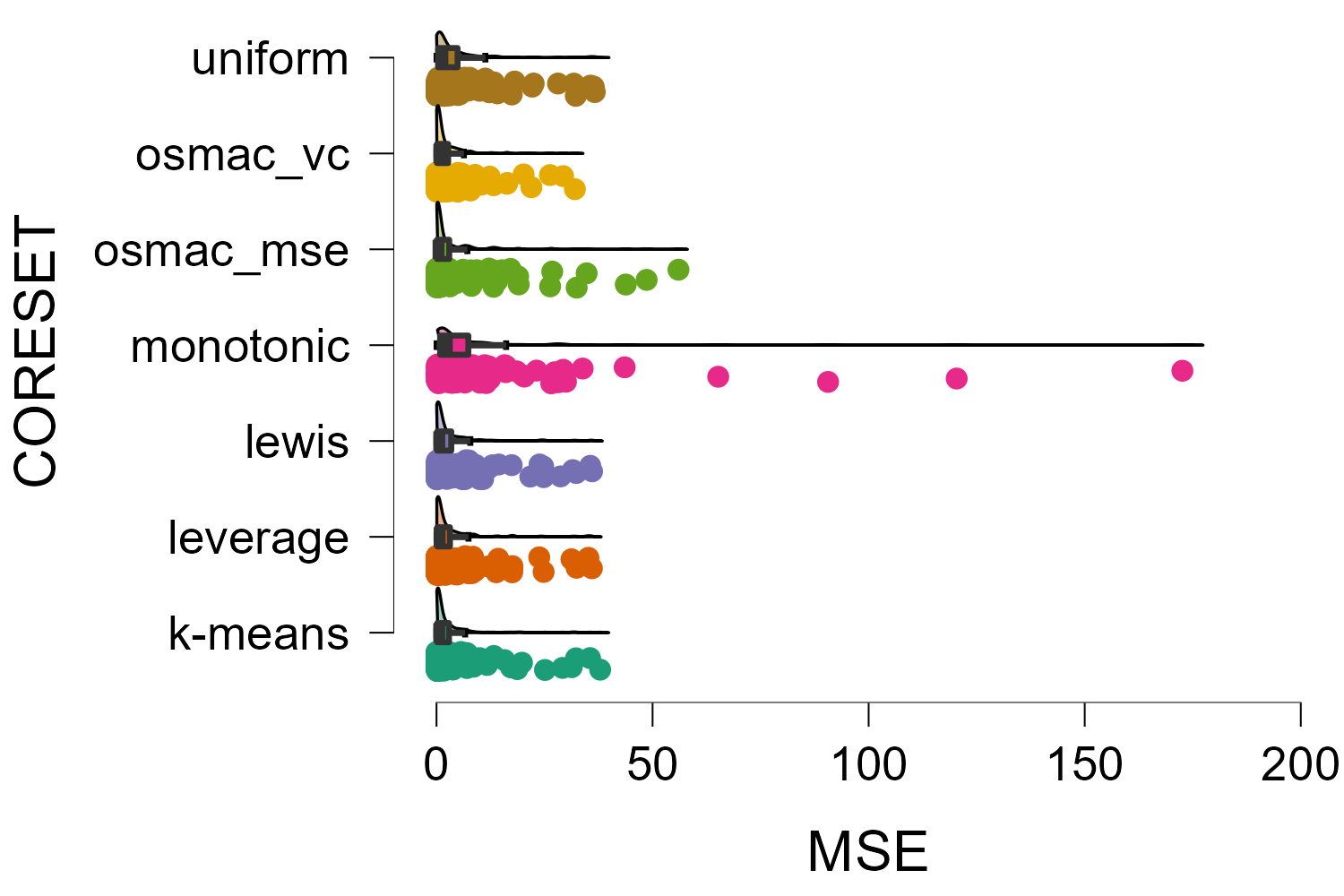}
    \includegraphics[width=\columnwidth, trim=0cm 0cm 0cm 0cm]{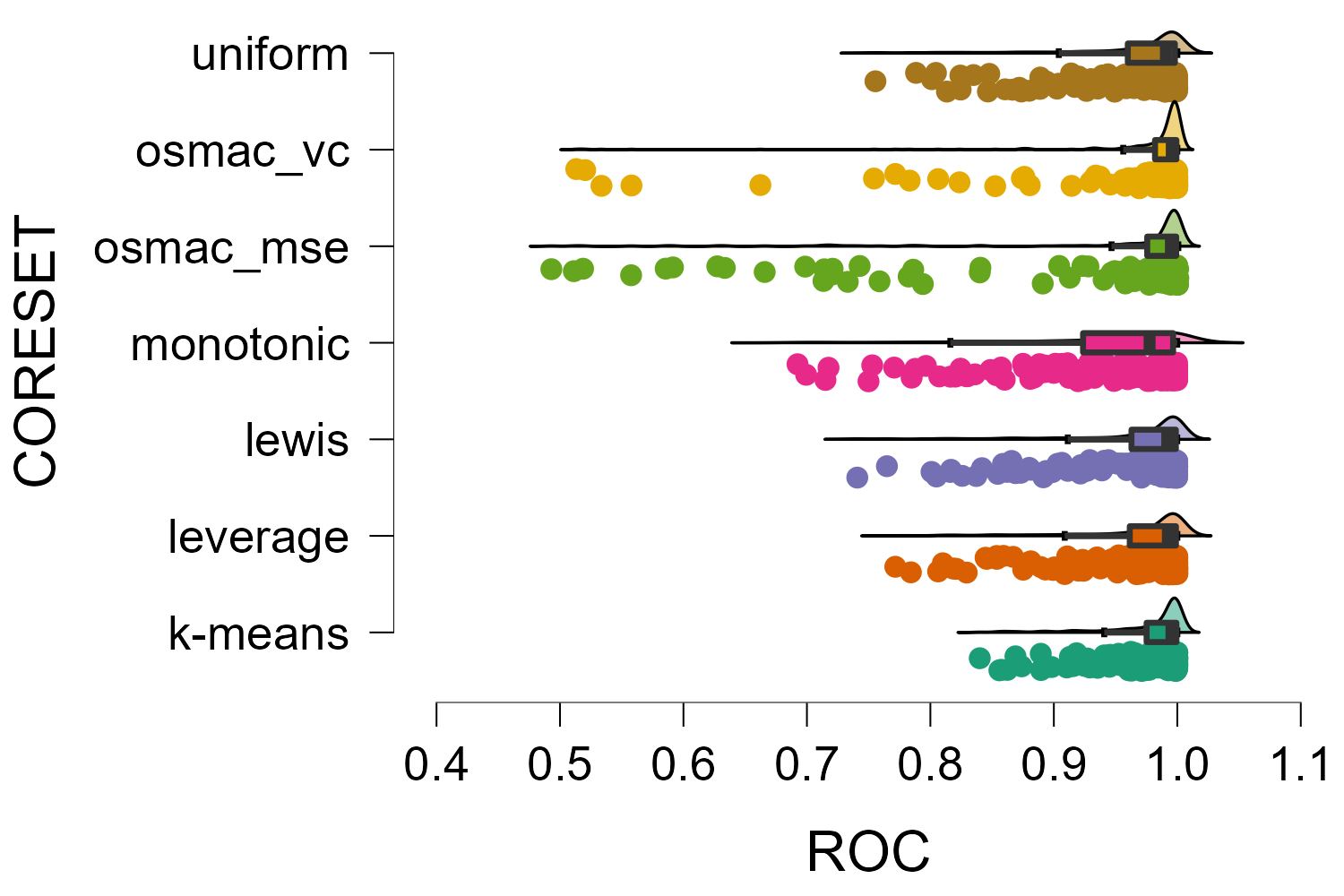}
\end{subfigure}
\caption{Comparing the distribution of scores between each method for all three metrics. In all cases the uniform baseline is as competitive as any other approach. We note the \texttt{Monotonic} method has worse outliers when measured by NLL or MSE, and that the \texttt{OSMAC} methods have a worse tail when measured by ROC.}
\label{fig:dist_cmp}
\end{figure}

Given the apparent failure of these methods to beat the naive baseline, it is natural to wonder how such a situation came to be. First, we remind the reader that in \autoref{fig:main} the behavior of the coreset algorithms varies from dataset to dataset, and our work is testing more datasets than prior works have done. By chance and happenstance, testing on fewer datasets may lead one to make conclusions of superior results, a common issue in most sciences today \cite{Ioannidis2005} and in machine learning competitions \cite{10.1145/502512.502576}. For example, Webb spam and KDD cup were two of the most widely used datasets in the coreset literature, and coreset methods happen to perform well on those.

Furthermore, most studies only included one or two baselines. It only takes one paper to draw conclusions of improvement against a baseline for later works to follow suit and exclude the ``beaten'' baseline in favor of newer methods. We highlighted an example of this where the $k$-means method turned out as one of the best performers but was never evaluated in later works. Through this lens it becomes clear how subsequent papers can, following reasonable practices of replicating prior methods as their baseline, result in an erroneous conclusion of improvement. This erroneous conclusion can still occur when even when testing on more datasets once a less ideal baseline model enters evaluation. 
Our experiment is in fact the first comprehensive comparison among several prior methods. As shown in \autoref{tbl:bases}, there may be a statistically significant difference between two methods while each has no statistical difference with the uniform random baseline. 

\begin{table}[!h]
\centering
\caption{Comparison of coreset methods with each other. In many cases one coreset method is significantly better than another, but not better than uniform random sampling as shown in \autoref{tbl:uniform}. This helps explain why many works have drawn conclusions of significance by comparing against other methods, but failing to include the competitive but naive baseline.} \label{tbl:bases}
\adjustbox{max width=\columnwidth}{%
\begin{tabular}{@{}lccc@{}}
\toprule
\multicolumn{1}{c}{Comparison} & NLL                      & $\beta$ MSE              & ROC                      \\ \midrule
\texttt{k-means} - \texttt{Leverage}             & 1                        & 1                        & 1                        \\
\texttt{k-means} - \texttt{Lewis}                & 1                        & 1                        & 1                        \\
\texttt{k-means} - \texttt{Monotonic}            & \textbf{\textless{}1e-4} & \textbf{\textless{}1e-4} & \textbf{\textless{}1e-4} \\
\texttt{k-means} - \texttt{OSMAC\_MSE}           & 1                        & 1                        & 1                        \\
\texttt{k-means} - \texttt{OSMAC\_vc}            & 1                        & 1                        & 1                        \\
\texttt{Leverage} - \texttt{Lewis}               & 1                        & 1                        & 1                        \\
\texttt{Leverage} - \texttt{Monotonic}           & \textbf{\textless{}1e-4} & \textbf{\textless{}1e-4} & \textbf{0.009}           \\
\texttt{Leverage} - \texttt{OSMAC\_MSE}          & 1                        & 1                        & 1                        \\
\texttt{Leverage} - \texttt{OSMAC\_vc}           & 1                        & 1                        & 1                        \\
\texttt{Lewis} - \texttt{Monotonic}              & \textbf{\textless{}1e-4} & \textbf{\textless{}1e-4} & 0.02                     \\
\texttt{Lewis} - \texttt{OSMAC\_MSE}             & 1                        & 1                        & 1                        \\
\texttt{Lewis} - \texttt{OSMAC\_vc}              & 1                        & 1                        & 0.645                    \\
\texttt{Monotonic} - \texttt{OSMAC\_MSE}         & \textbf{\textless{}1e-4} & \textbf{\textless{}1e-4} & \textbf{\textless{}1e-4} \\
\texttt{Monotonic} - \texttt{OSMAC\_vc}          & \textbf{\textless{}1e-4} & \textbf{\textless{}1e-4} & \textbf{\textless{}1e-4} \\
\texttt{OSMAC\_MSE} - \texttt{OSMAC\_vc}         & 1                        & 1                        & 1                        \\ \bottomrule
\end{tabular}
}
\end{table}

Finally one may ask if the overall performance does not differ, does the distribution of the performances differ? We show that they do in \autoref{fig:dist_cmp}, but that the uniform baseline is still competitive in all cases. Instead we see that the \texttt{Monotonic} and \texttt{OSMAC} methods have poorly behaved outliers in their distribution of scores, depending on the method of evaluation being used. From visual inspection, we note that $k$-means, \texttt{Leverage}, and \texttt{Lewis} appear the most effective of the coreset methods.

\subsection{Effect of regularization}

As discussed in \autoref{sec:methods}, regularization impacts the effectiveness of coreset methods in logistic regression. In particular, although results from \cite{munteanu2018coresets} indicate that the availability of coresets depends on the complexity of the data, the uniform baseline is known to be effective on all datasets when the model is regularized \cite{curtin2019coresets}. We run a followup experiment where we vary the L2 penalty of the classifiers beyond that of our main experiment. Our findings in \autoref{fig:l2} confirm experimentally that increasing regularization makes the gap between uniform sampling and the best-performing methods vanish. 

One may ask whether further decreasing regularization strength in our main experiment would impact our findings by weakening the uniform subsample. We replicate our experiment with $\lambda=10^{-7}$ with results in  \autoref{tbl:uniform_reg} and Appendix \autoref{tbl:bases_reg}. We find that the situation becomes worse, and no method outperforms uniform random sampling. Further lowering $\lambda$ hardly changed the regression coefficients, so we expect similar results for any $\lambda<10^{-7}$. Similarly, as we showed earlier in accordance with recent theoretical work that uniform sampling only improves with stronger regularization, we conclude that our results are likely to hold true over all $\lambda$.

\begin{table}[h]
\caption{Comparisons with uniform baseline for logistic regression coresets, with multiple test corrected $p$-value in the three right most columns. Regularization is further weakened to $\lambda=10^{-7}$ compared to the main experiment. Note the final column shows that \texttt{OSMAC\_vc} is significant \textit{worse than uniform}, so none of the methods are statistically improving over the naive baseline.} \label{tbl:uniform_reg}
\centering
\begin{tabular}{@{}lccc@{}}
\toprule
\multicolumn{1}{c}{Comparison} & NLL   & $\beta$ MSE & ROC   \\ \midrule
\texttt{k-means} - uniform              & 0.140 & 0.482       & 0.288 \\
\texttt{Leverage} - uniform             & 1.000 & 0.733       & 1.000 \\
\texttt{Lewis} - uniform                & 1.000 & 0.18        & 1.000 \\
\texttt{Monotonic} - uniform            & 0.215 & 1.000       & 0.548 \\
\texttt{OSMAC\_MSE} - uniform           & 1.000 & 0.433       & 1.000 \\
\texttt{OSMAC\_vc} - uniform            & 1.000 & 0.152       & 1.000 \\ \bottomrule
\end{tabular}
\end{table}

\section{Conclusions} \label{sec:conclusion}

In this work we have performed a thorough evaluation against many seminal methods used in the coreset and sub-sampling literature, used with respect to logistic regression. When compared against a naive baseline of uniform random selection of a subset, and measured by three possible metrics of improvement, we find that almost none of these classic approaches improves upon the naive baseline. Our results call into question the need for larger diversity of evaluation sets and benchmarks to be used in coreset research. 

\bibliography{aaai23.bib}

\clearpage
\appendix
\onecolumn

\input{appendix}

\end{document}

%% file: appendix.tex
\section{Remaining figures for main experiment}

\begin{figure*}[h]
\centering
\begin{subfigure}{0.99\textwidth}
\centering
\hspace*{\fill}  Relative NLL error $(\downarrow)$ \hfill Coefficient MSE $(\downarrow)$ \hfill Relative ROC $(\uparrow)$ \hspace*{\fill}
\end{subfigure}
\begin{subfigure}{0.99\textwidth}
\begin{turn}{90} 
\phantom{BufferBuffer}chemreact
\end{turn}
    \centering
    \hspace*{\fill}
    \includegraphics[width=0.29\textwidth, trim=0cm 0cm 0cm 0cm]{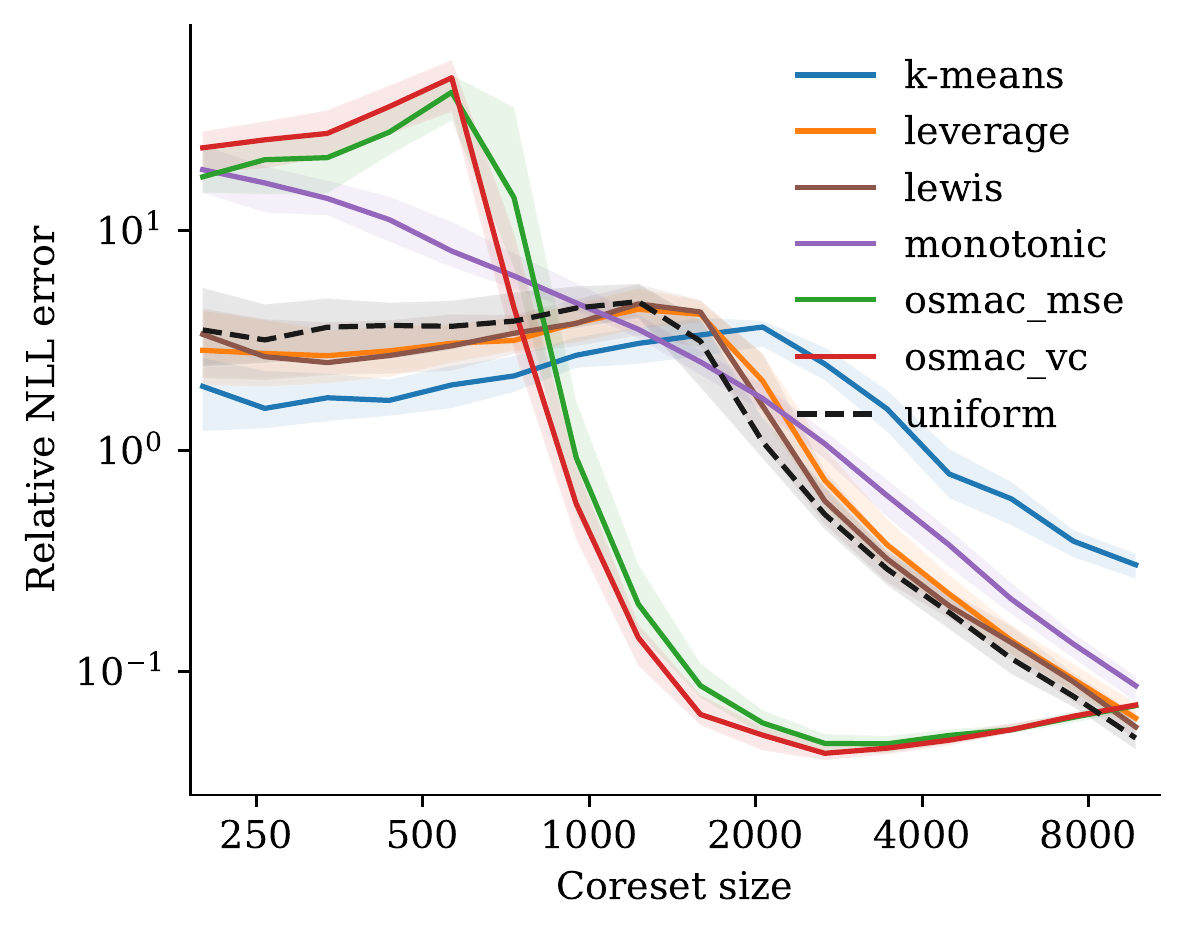}
    \hfill
    \includegraphics[width=0.29\textwidth, trim=0cm 0cm 0cm 0cm]{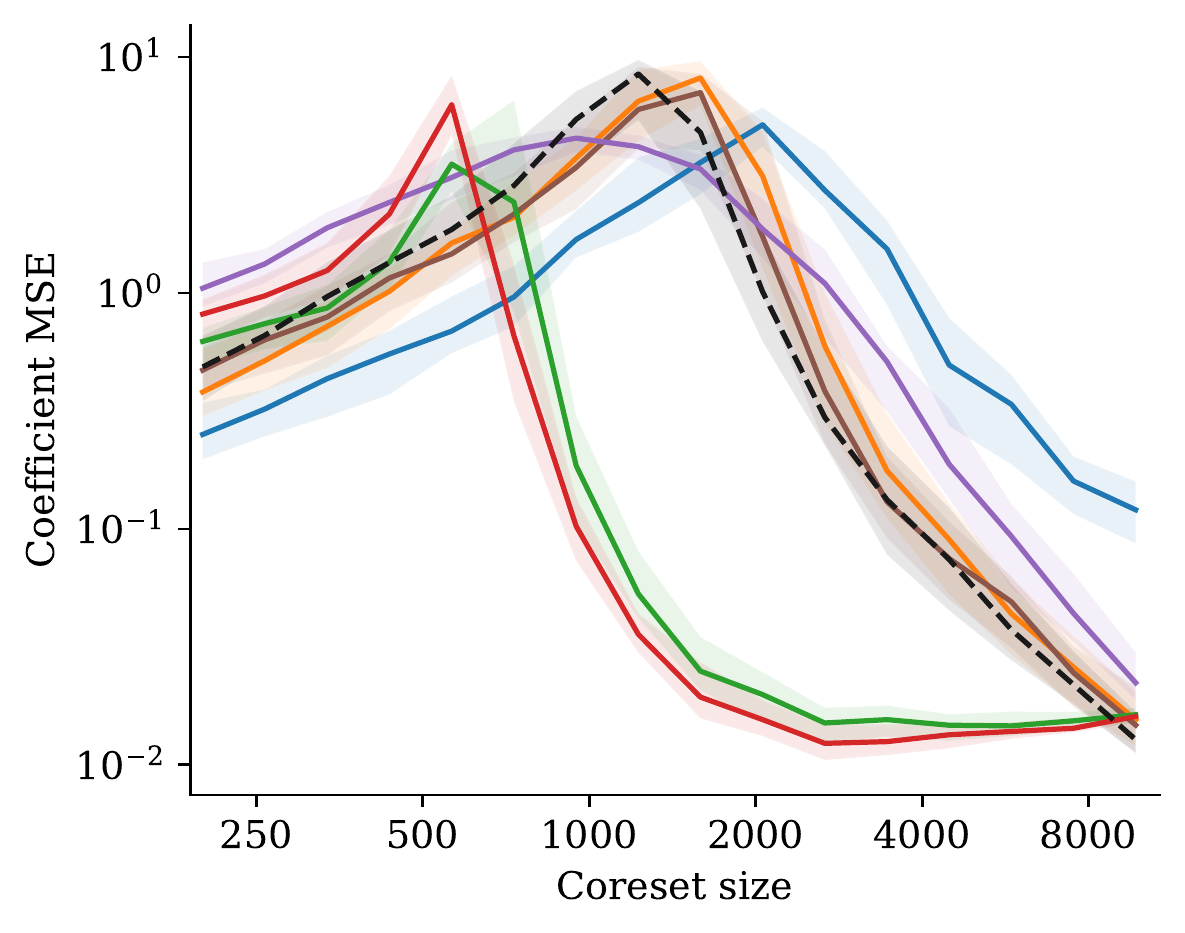}
    \hfill
    \includegraphics[width=0.29\textwidth, trim=0cm 0cm 0cm 0cm]{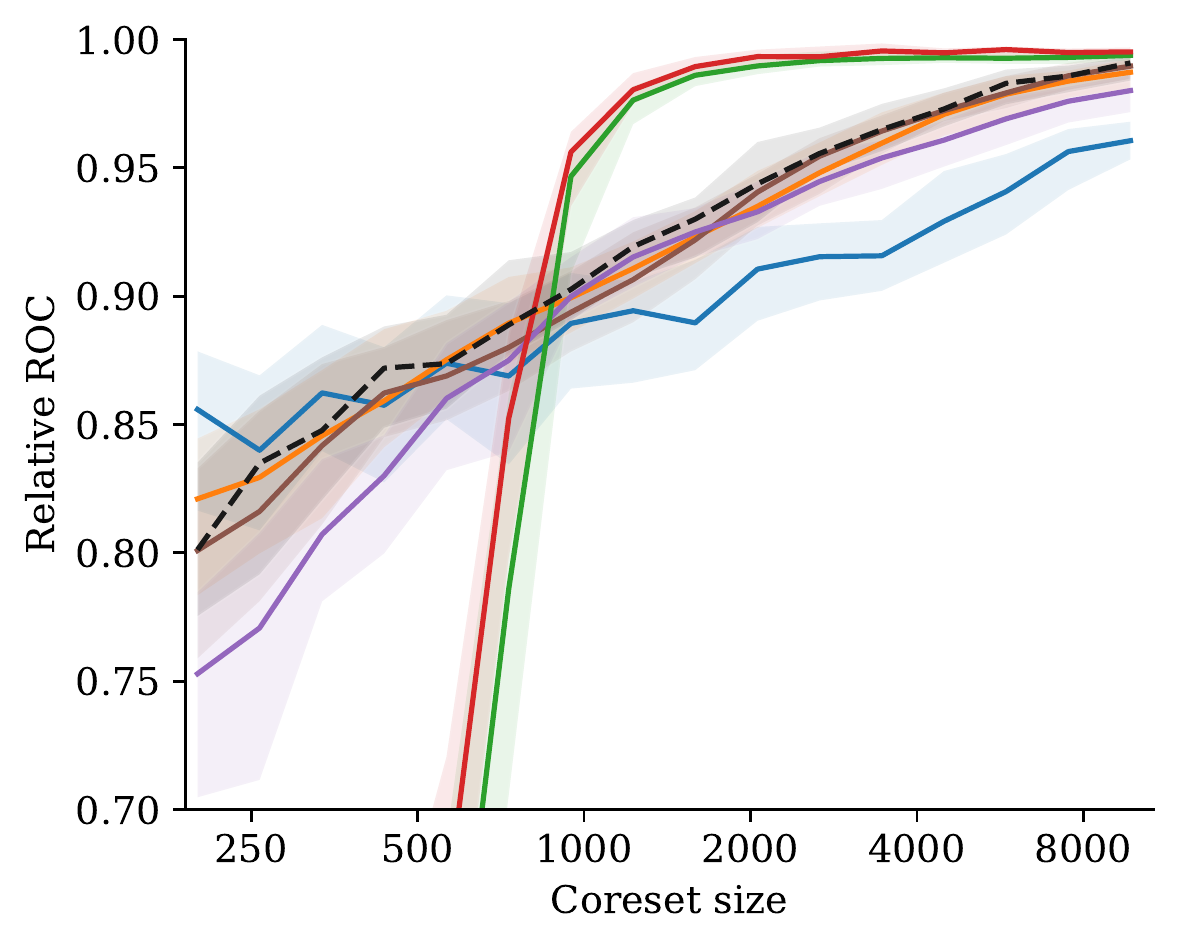}
    \hspace*{\fill}
\end{subfigure}

\begin{subfigure}{0.99\textwidth}
\begin{turn}{90} 
\phantom{BufferBuffer}census
\end{turn}
    \centering
    \hspace*{\fill}
    \includegraphics[width=0.29\textwidth, trim=0cm 0cm 0cm 0cm]{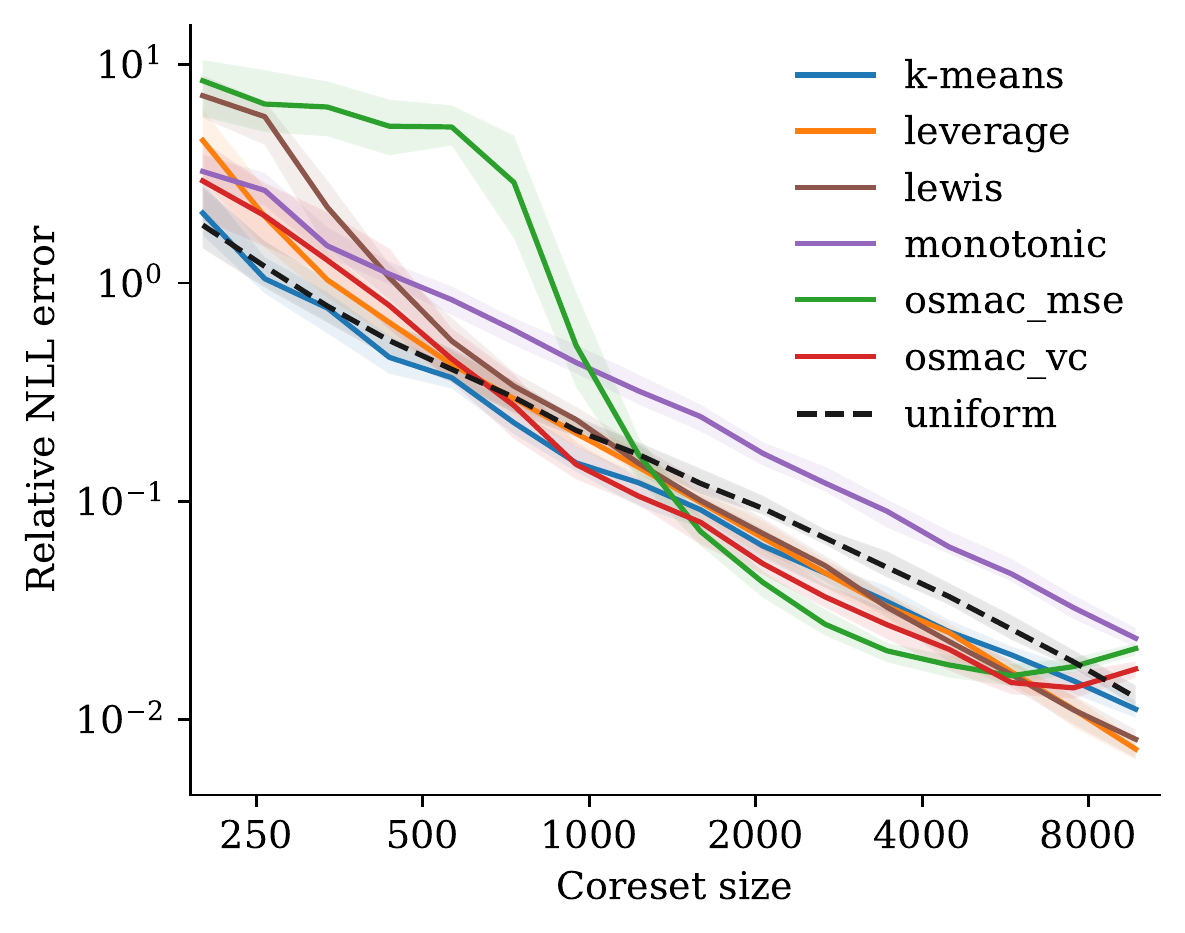}
    \hfill
    \includegraphics[width=0.29\textwidth, trim=0cm 0cm 0cm 0cm]{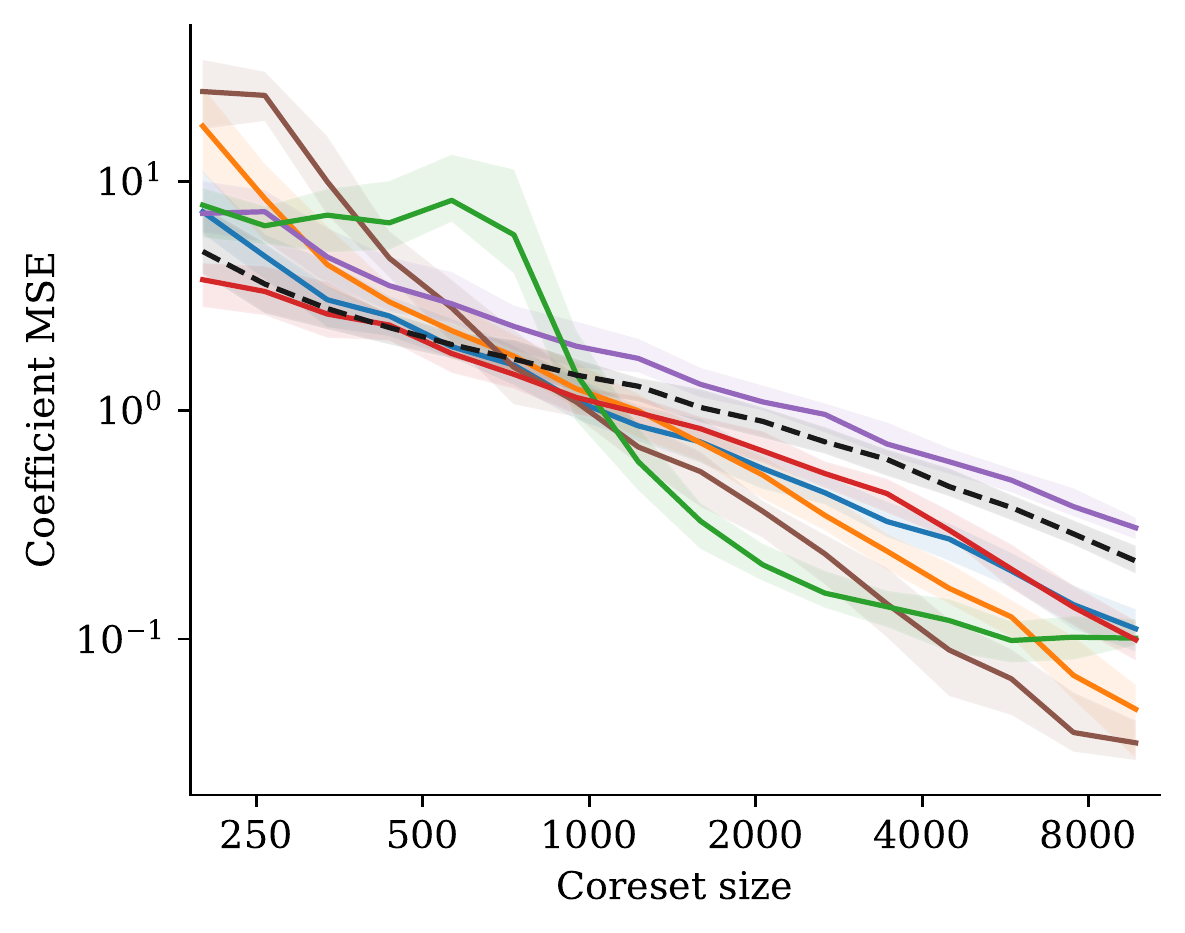}
    \hfill
    \includegraphics[width=0.29\textwidth, trim=0cm 0cm 0cm 0cm]{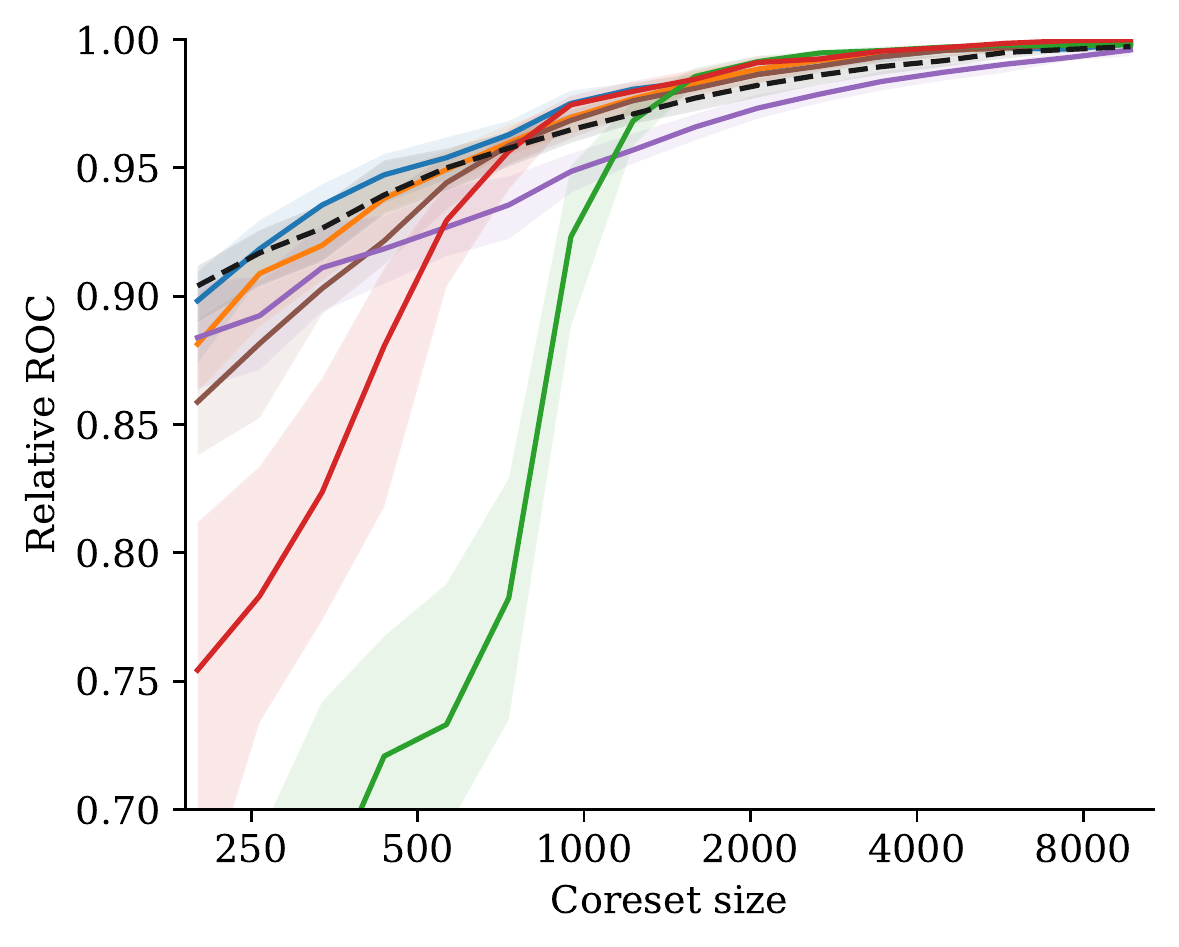}
    \hspace*{\fill}
\end{subfigure}

\begin{subfigure}{0.99\textwidth}
\begin{turn}{90} 
\phantom{BufferBuffer}bank
\end{turn}
    \centering
    \hspace*{\fill}
    \includegraphics[width=0.29\textwidth, trim=0cm 0cm 0cm 0cm]{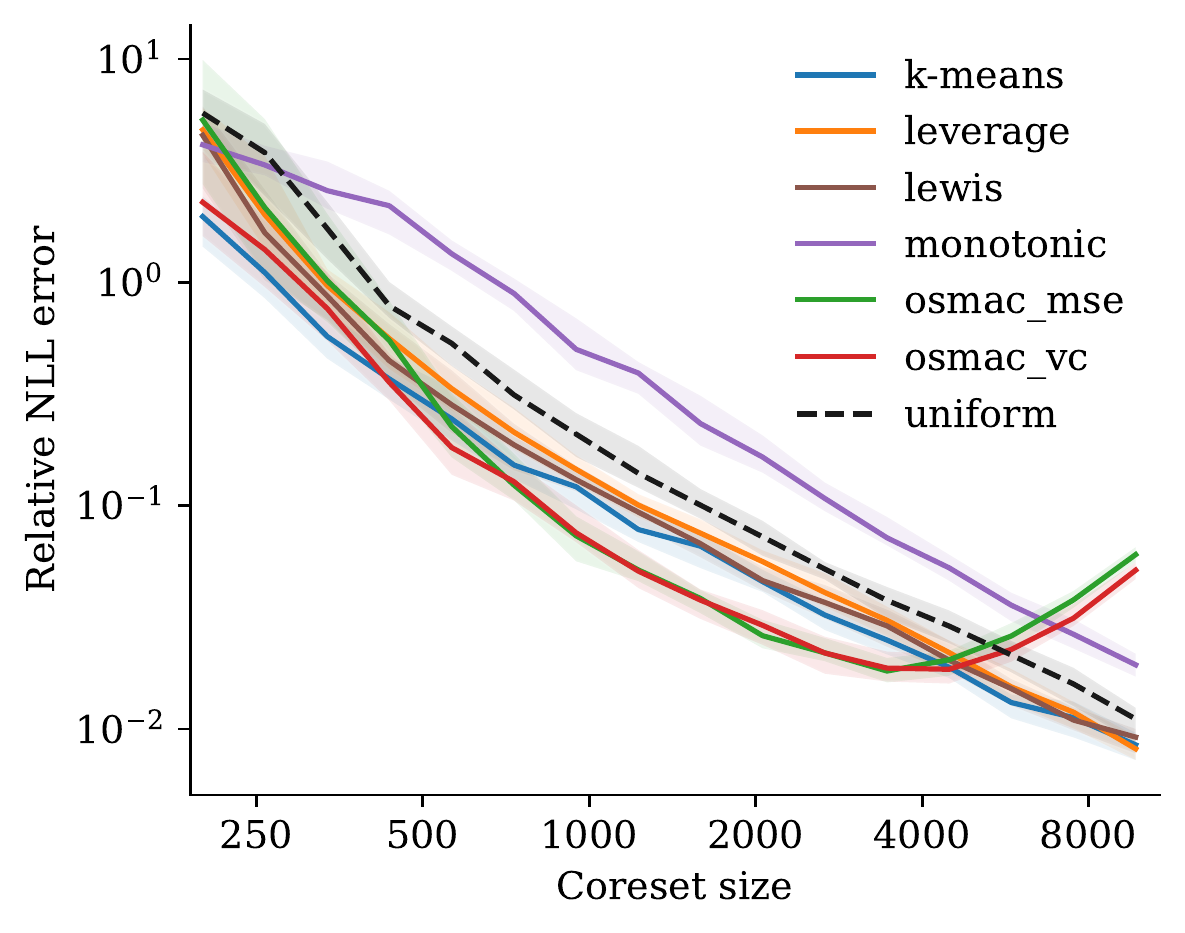}
    \hfill
    \includegraphics[width=0.29\textwidth, trim=0cm 0cm 0cm 0cm]{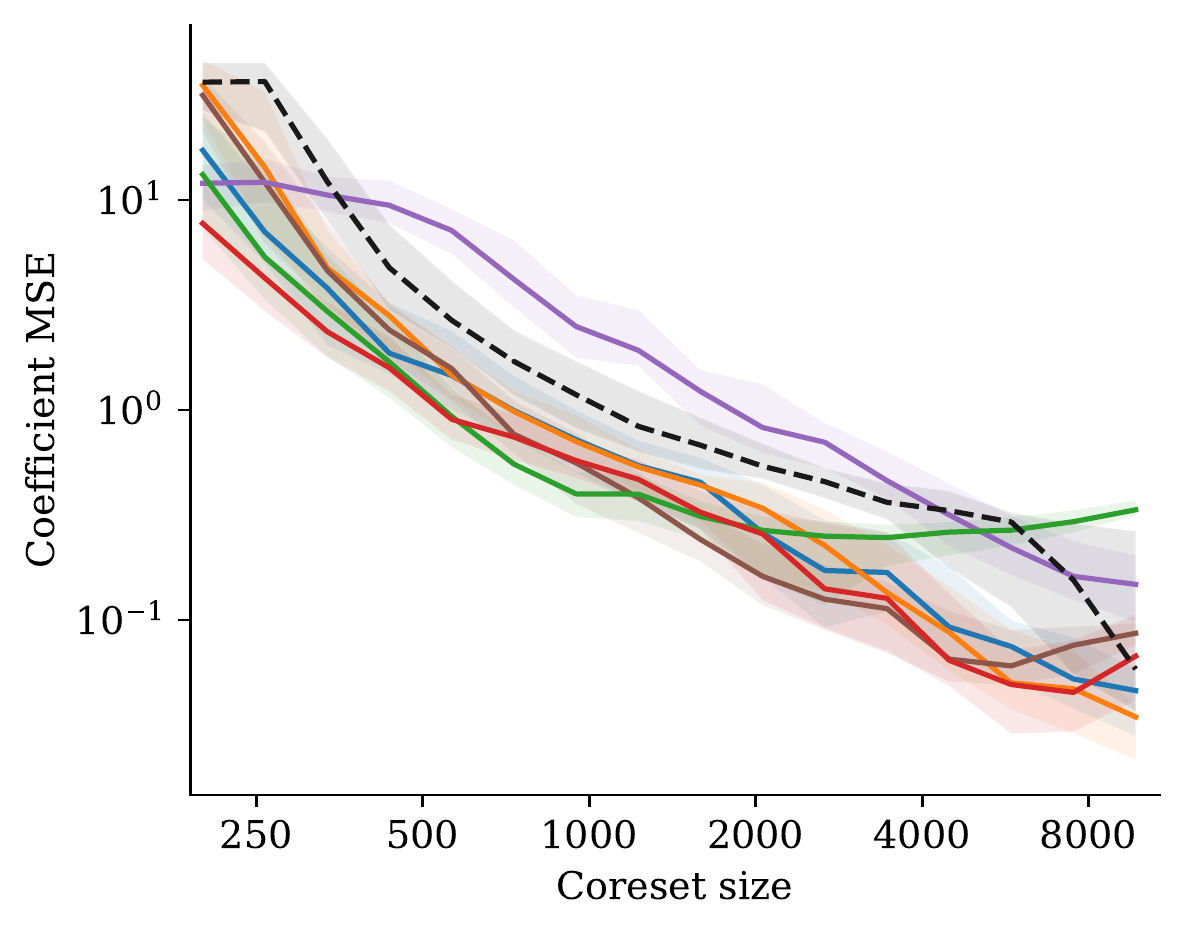}
    \hfill
    \includegraphics[width=0.29\textwidth, trim=0cm 0cm 0cm 0cm]{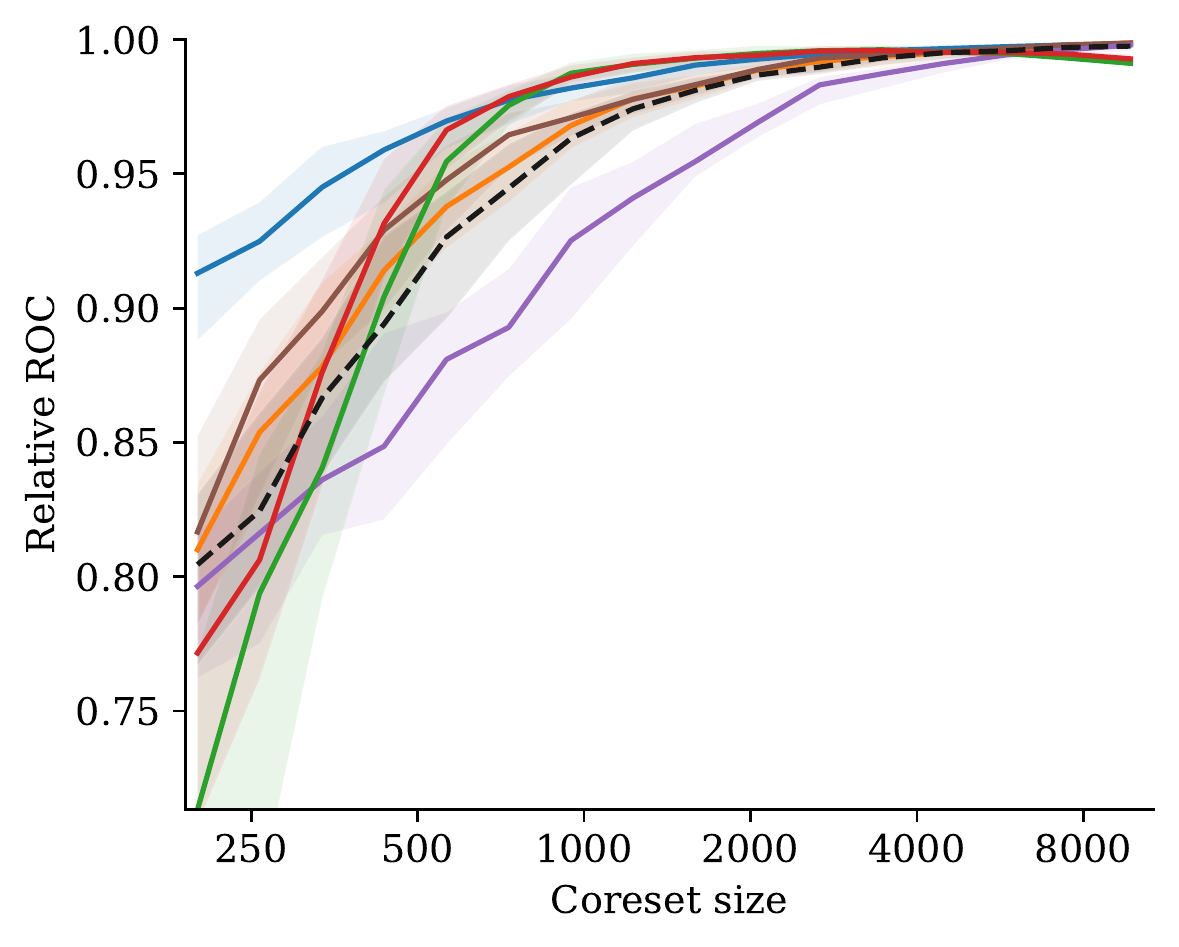}
    \hspace*{\fill}
\end{subfigure}

\caption{All subsampling methods are compared over a range of 25 subsample sizes. Each row represents one dataset and each column represents a different evaluation metric. Each point represents a median of 50 replications, with shaded areas representing interquartile range. The three smaller datasets are shown here, with the remaining in the main paper.}
\label{fig:main_more}
\end{figure*}

\FloatBarrier
\clearpage
\section{Effect of L2 regularization}

\begin{figure*}[h]
\centering
\begin{subfigure}{0.99\textwidth}
\centering
\hspace*{\fill}  $\lambda=10^{-3}$ \hfill $\lambda=10^{-1}$ \hfill $\lambda=10$ \hspace*{\fill}
\end{subfigure}
\begin{subfigure}{0.99\textwidth}
\begin{turn}{90} 
\phantom{BufferBuffer}KDD cup
\end{turn}
    \centering
    \hspace*{\fill}
    \includegraphics[width=0.29\textwidth, trim=0cm 0cm 0cm 0cm]{figs/l2_exp/fig_kddcup_rel_nll_error_0.001.pdf}
    \hfill
    \includegraphics[width=0.29\textwidth, trim=0cm 0cm 0cm 0cm]{figs/l2_exp/fig_kddcup_rel_nll_error_0.1.pdf}
    \hfill
    \includegraphics[width=0.29\textwidth, trim=0cm 0cm 0cm 0cm]{figs/l2_exp/fig_kddcup_rel_nll_error_10.0.pdf}
    \hspace*{\fill}
\end{subfigure}

\begin{subfigure}{0.99\textwidth}
\begin{turn}{90} 
\phantom{BufferBuffer}Webb spam
\end{turn}
    \centering
    \hspace*{\fill}
    \includegraphics[width=0.29\textwidth, trim=0cm 0cm 0cm 0cm]{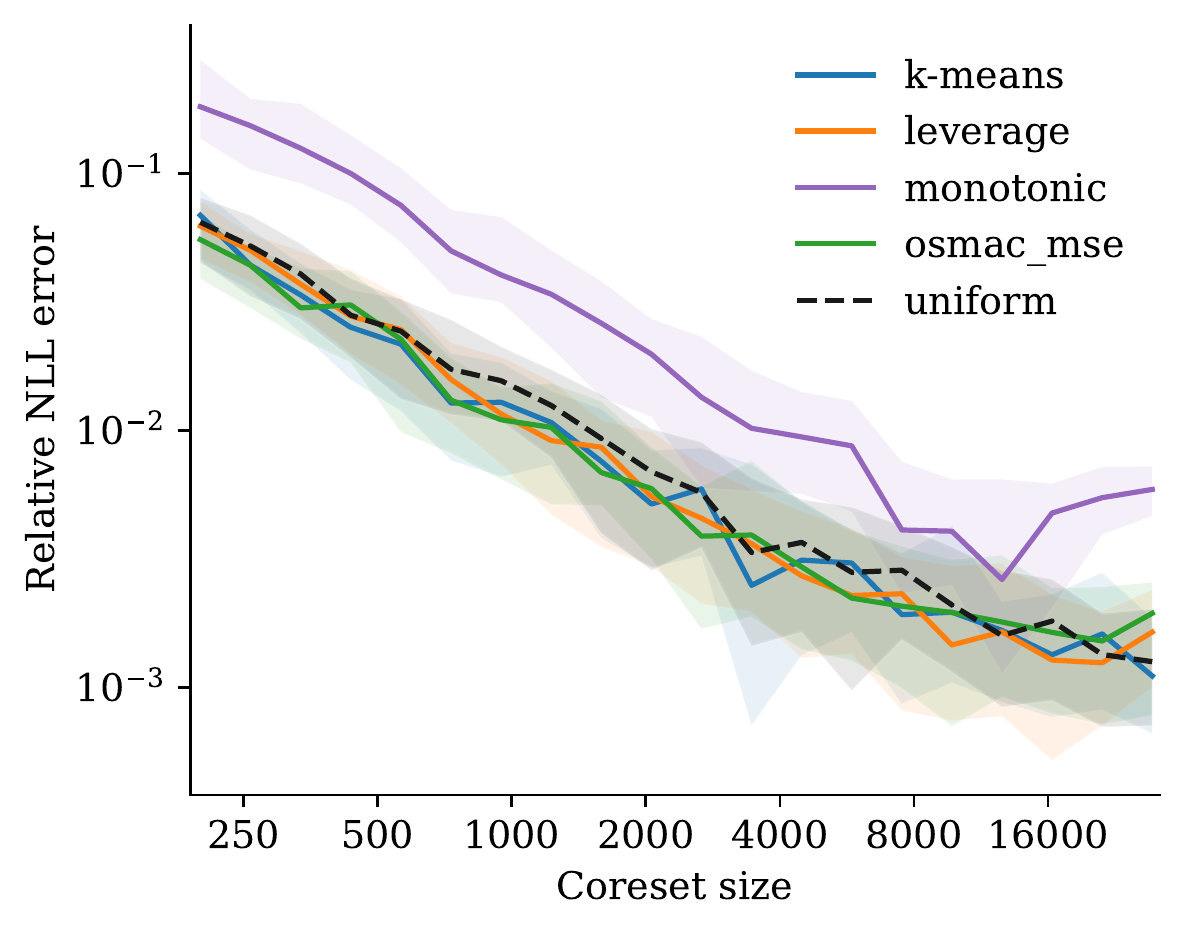}
    \hfill
    \includegraphics[width=0.29\textwidth, trim=0cm 0cm 0cm 0cm]{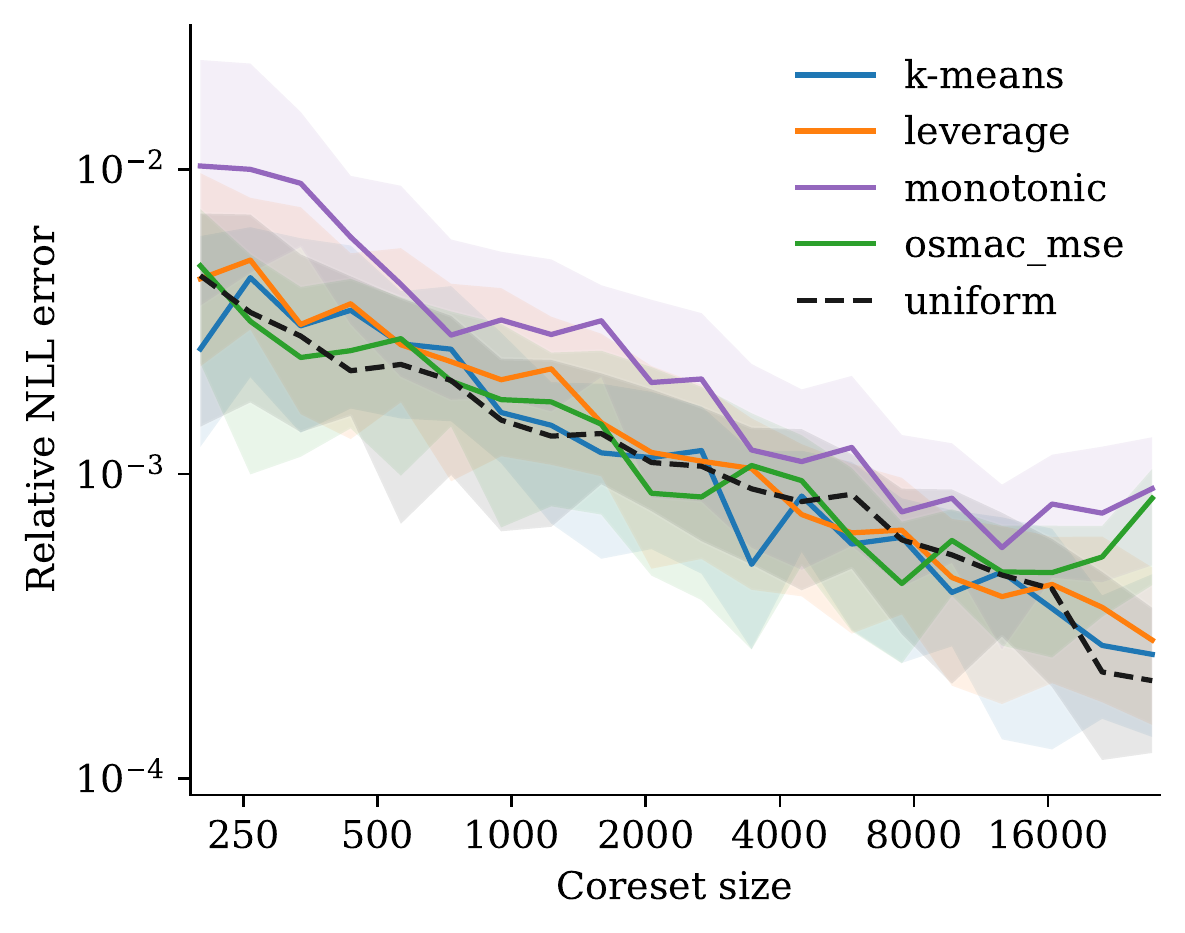}
    \hfill
    \includegraphics[width=0.29\textwidth, trim=0cm 0cm 0cm 0cm]{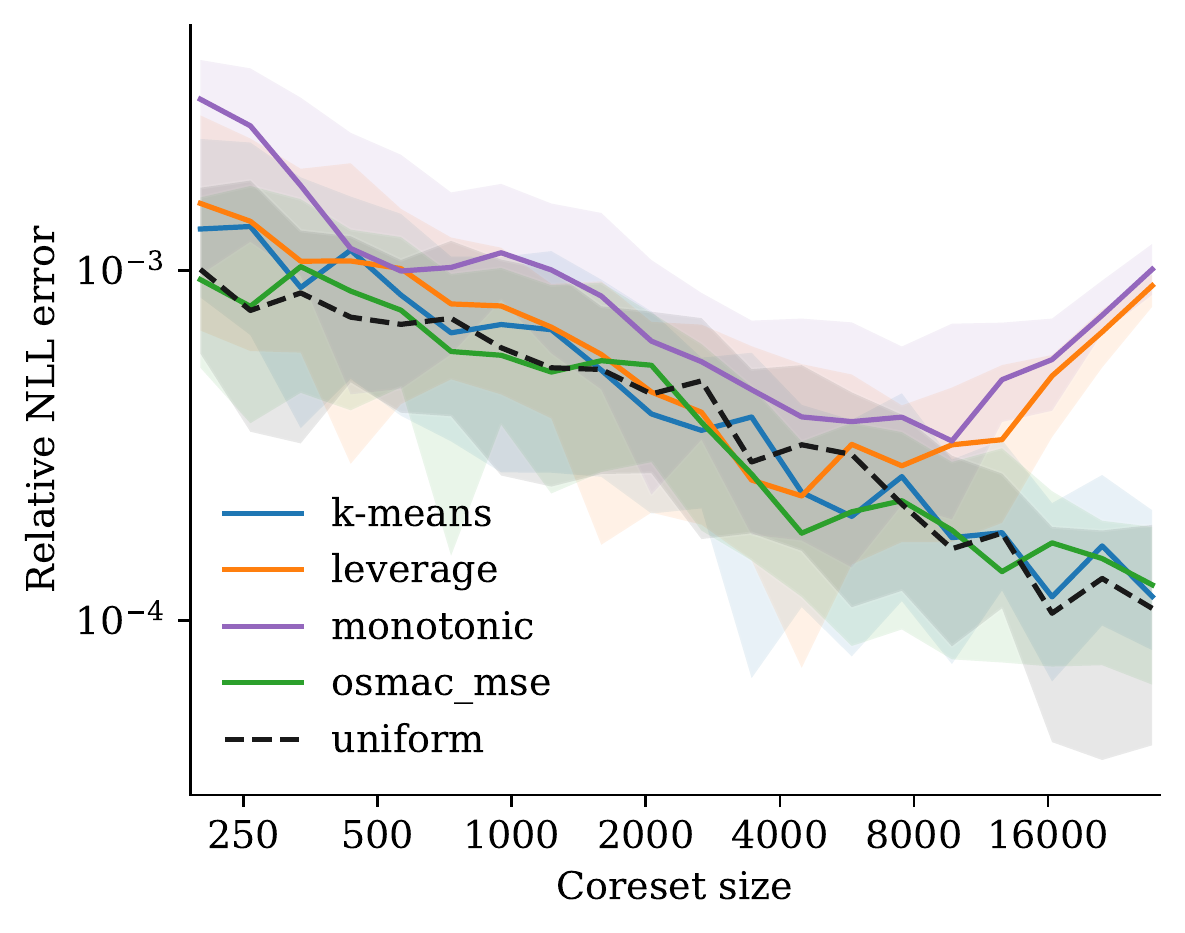}
    \hspace*{\fill}
\end{subfigure}

\begin{subfigure}{0.99\textwidth}
\begin{turn}{90} 
\phantom{BufferBuffer}covertype
\end{turn}
    \centering
    \hspace*{\fill}
    \includegraphics[width=0.29\textwidth, trim=0cm 0cm 0cm 0cm]{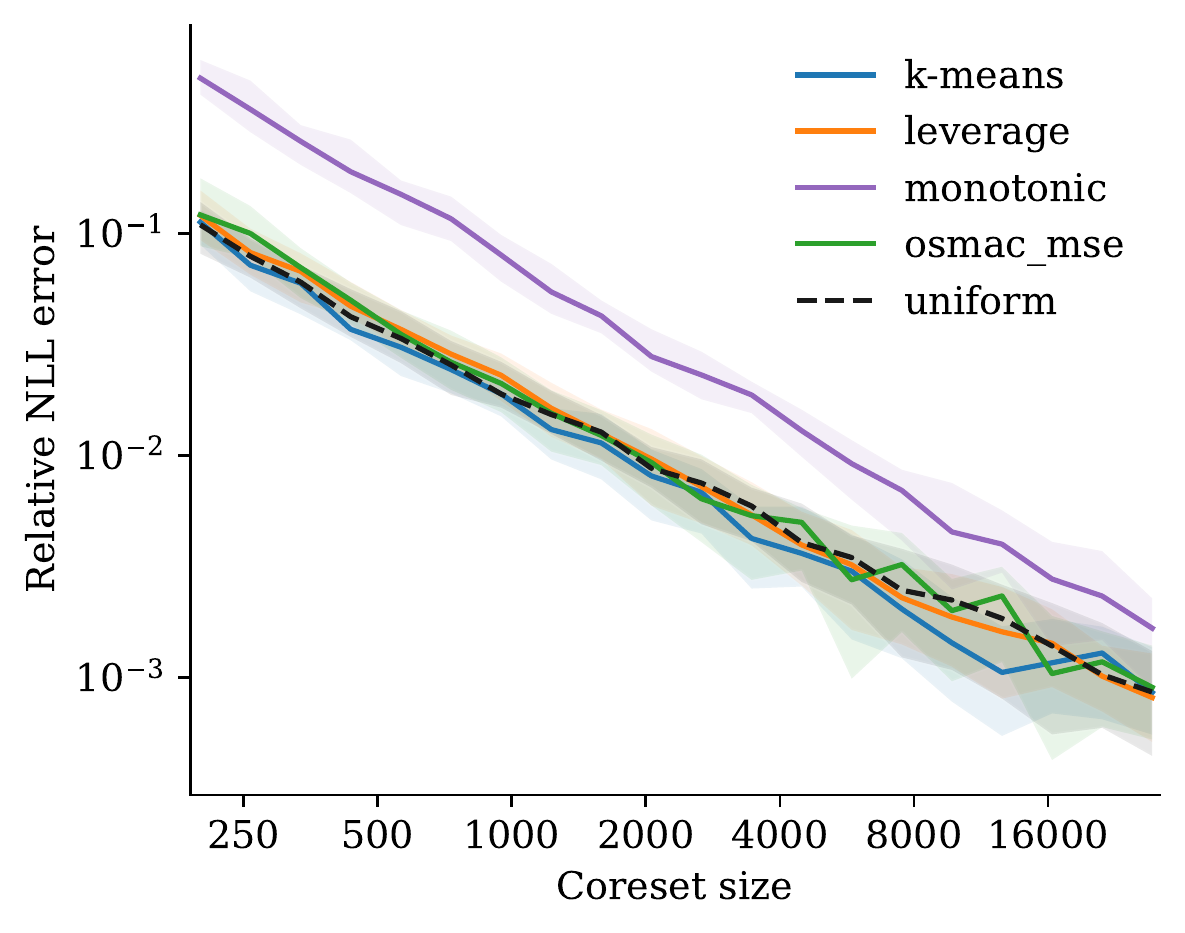}
    \hfill
    \includegraphics[width=0.29\textwidth, trim=0cm 0cm 0cm 0cm]{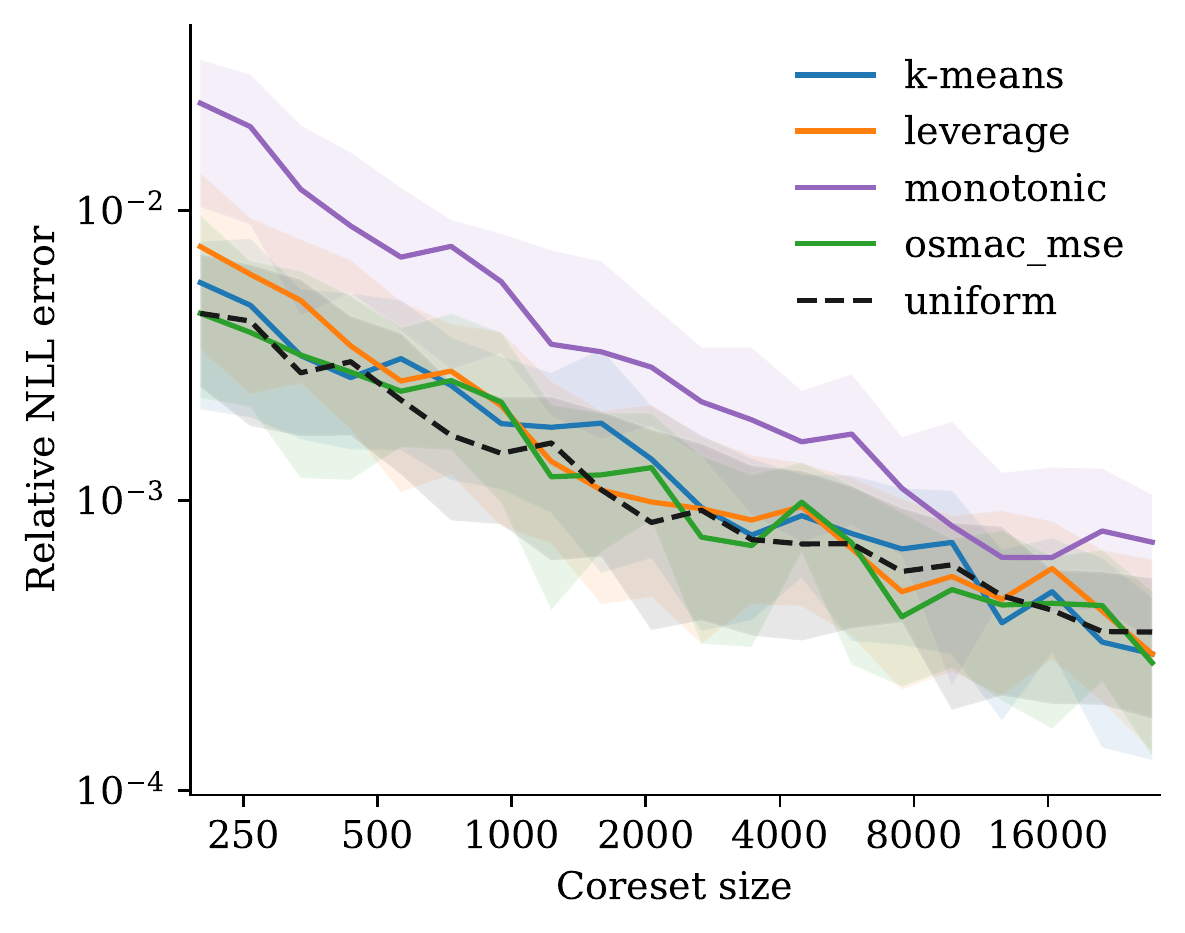}
    \hfill
    \includegraphics[width=0.29\textwidth, trim=0cm 0cm 0cm 0cm]{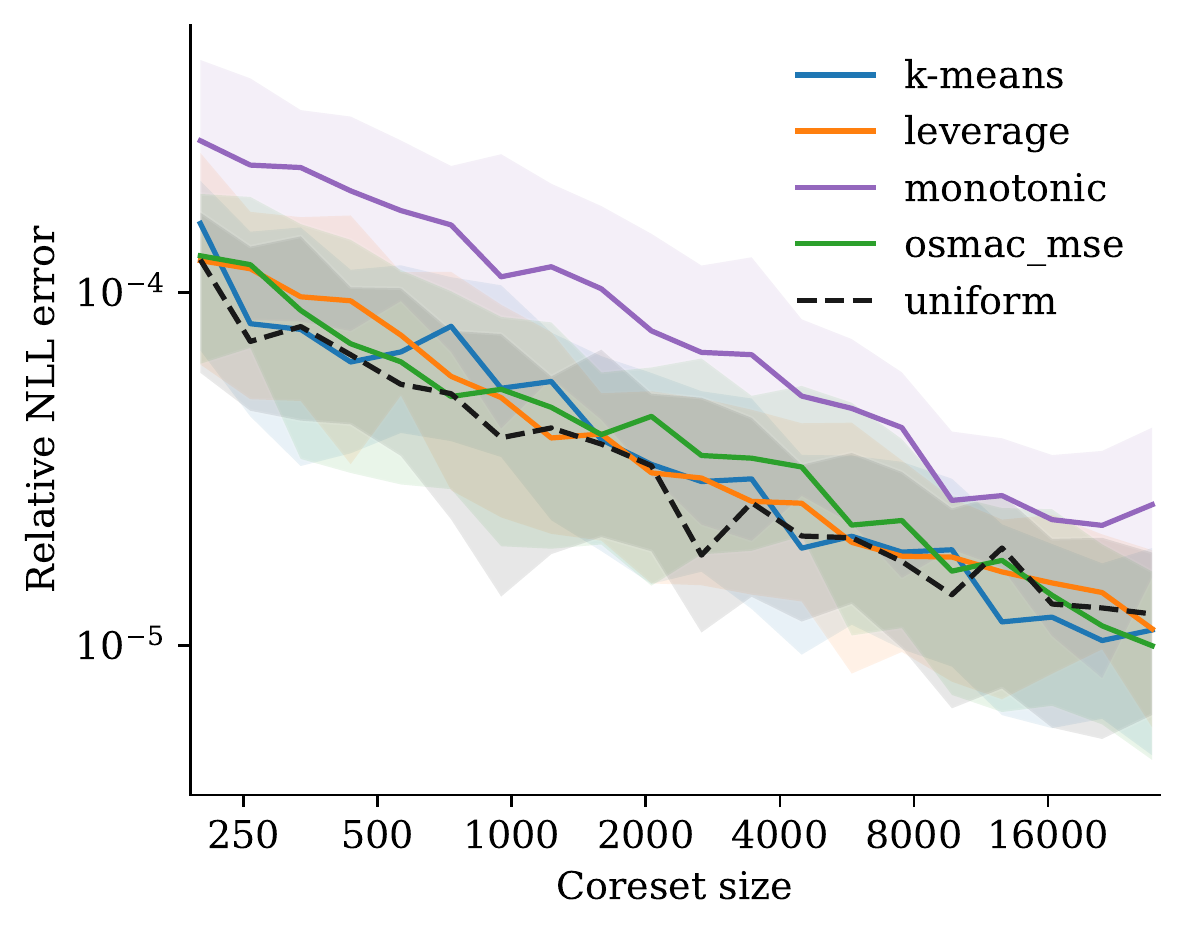}
    \hspace*{\fill}
\end{subfigure}

\caption{Additional results. Increasing L2 regularization noticeably improves the efficiency of uniform subsampling relative to other strong-performing methods. This shows concordance with the theoretical results in \cite{curtin2019coresets}.}
\label{fig:l2_more}
\end{figure*}

\FloatBarrier
\clearpage
\section{Replicated analysis for weakened regularization}

The final table was not shown in main paper due to space, and is a comparison among coreset methods with regularization $\lambda=10^{-7}$.

\begin{table}[h]
\centering
\caption{Comparison of coreset methods with each other for logistic regression with further weakened regularization. In many cases one coreset method is significantly better than another, but not better than uniform random sampling as shown in \autoref{tbl:uniform_reg}. } \label{tbl:bases_reg}
\adjustbox{max width=\columnwidth}{%
\begin{tabular}{@{}llll@{}}
\toprule
\multicolumn{1}{c}{Comparison} & \multicolumn{1}{c}{NLL} & \multicolumn{1}{c}{MSE} & \multicolumn{1}{c}{ROC} \\ \midrule
k-means - leverage             & 1                       & 1                       & 1                       \\
k-means - lewis                & 1                       & 1                       & 1                       \\
k-means - monotonic            & \textbf{2.710E-6}       & \textbf{4.969E-4}       & \textbf{5.764E-5}       \\
k-means - osmac\_mse           & 0.769                   & 1                       & 1                       \\
k-means - osmac\_vc            & 1                       & 1                       & 1                       \\
leverage - lewis               & 1                       & 1                       & 1                       \\
leverage - monotonic           & \textbf{0.001}          & \textbf{0.001}          & \textbf{0.013}          \\
leverage - osmac\_mse          & 1                       & 1                       & 1                       \\
leverage - osmac\_vc           & 1                       & 1                       & 1                       \\
lewis - monotonic              & \textbf{5.800E-4}       & \textbf{9.681E-5}       & \textbf{0.016}          \\
lewis - osmac\_mse             & 1                       & 1                       & 1                       \\
lewis - osmac\_vc              & 1                       & 1                       & 1                       \\
monotonic - osmac\_mse         & 0.03                    & \textbf{4.139E-4}       & 0.09                    \\
monotonic - osmac\_vc          & \textbf{0.002}          & \textbf{7.351E-5}       & \textbf{0.008}          \\
osmac\_mse - osmac\_vc         & 1                       & 1                       & 1                       \\ \bottomrule
\end{tabular}
}
\end{table}

\FloatBarrier
\clearpage
\section{L1 regularization}

One may further ask how the subsampling methods perform with L1 regularization, which induces sparsity in the estimated coefficients. To study this question, we selected the three datasets with most features $d$, which are census, chemreact, and Webb spam, and repeated our experiment with the L1 penalty equal to $\lambda=0.001$. We measured the support accuracy, which refers to the average agreement between $\hat\beta_C$ and $\hat\beta_{MSE}$ on whether a feature should be in the model. All methods improve as the coreset size increases, and methods like osmac\_mse and leverage appear to be effective performers. However, it appears that feature selection with high accuracy requires subsampling at the high end of our tested range. As all prior theory for subsampling target quantities such as the coefficient MSE which are agnostic to sparsity, we think this area is of interest for future study.

\begin{figure*}[h]
\centering
\begin{subfigure}{0.99\textwidth}
\centering
\hspace*{\fill}  census \hfill chemreact \hfill Webb spam \hspace*{\fill}
\end{subfigure}
\begin{subfigure}{0.99\textwidth}
\begin{turn}{90} 
\end{turn}
    \centering
    \hspace*{\fill}
    \includegraphics[width=0.29\textwidth, trim=0cm 0cm 0cm 0cm]{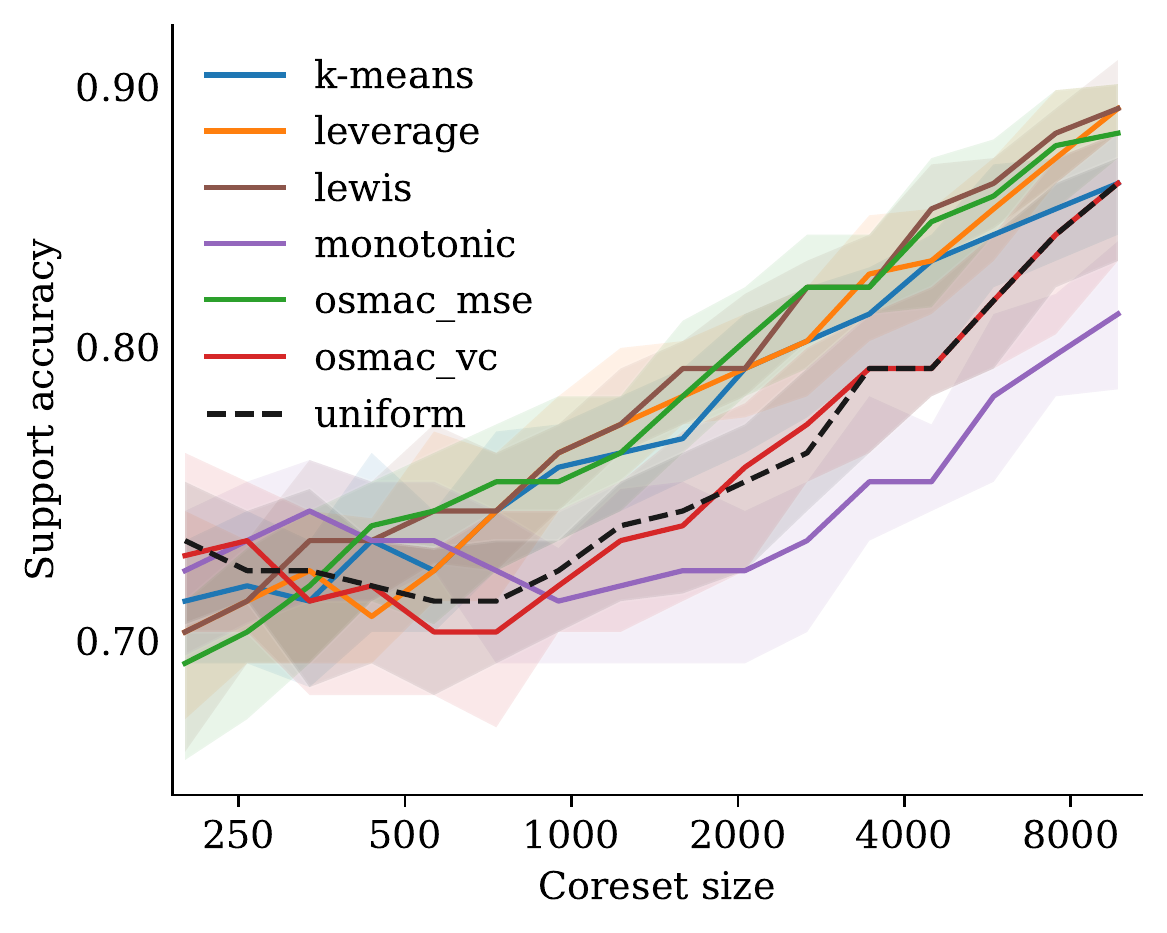}
    \hfill
    \includegraphics[width=0.29\textwidth, trim=0cm 0cm 0cm 0cm]{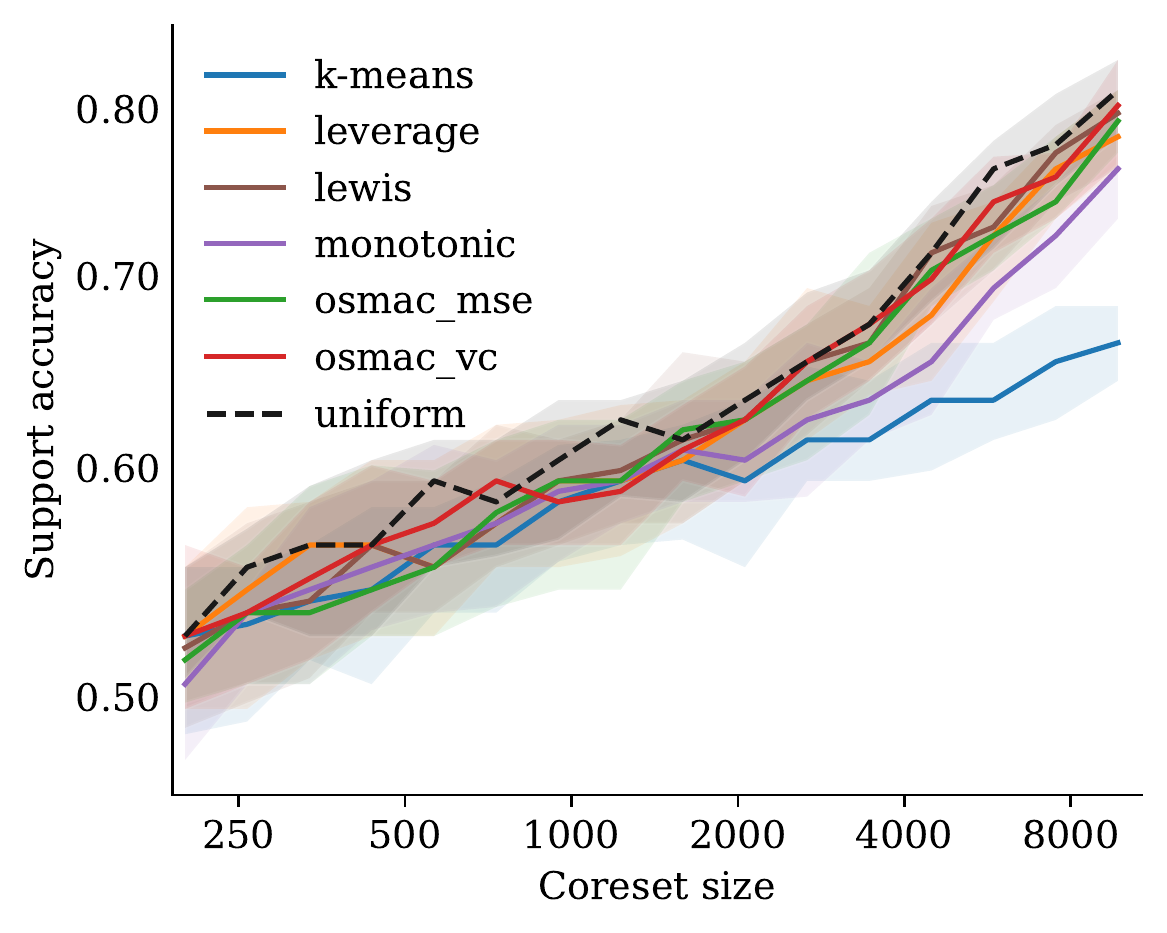}
    \hfill
    \includegraphics[width=0.29\textwidth, trim=0cm 0cm 0cm 0cm]{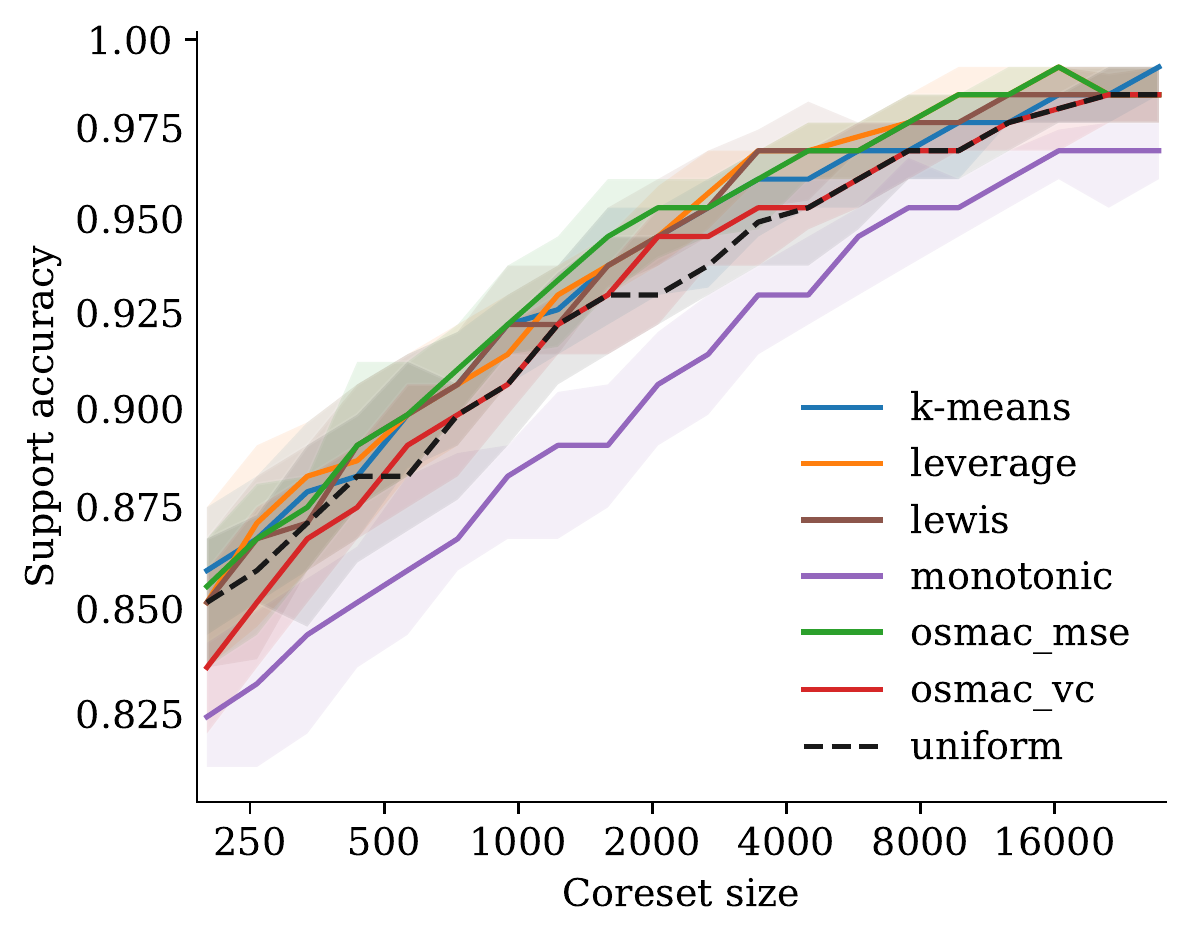}
    \hspace*{\fill}
\end{subfigure}

\caption{Support accuracy measures the average agreement between $\hat\beta_C$ and the $\hat\beta_{MSE}$ on whether a feature should be in the model, when L1 regularization is used to promote sparsity. The three datasets with the most features are shown here. $\lambda=0.001$ is used in all three figures. Higher is better.}
\label{fig:l1}
\end{figure*}

\FloatBarrier
\clearpage
\section{Experiment details}

\subsection{Datasets}

We detail our dataset sources and preprocessing steps here and in our code. Most datasets have previously been used in coreset papers, which was our primary motivation for including them. However, we include more features where possible, including categorical. In many cases, prior works only used a few numerical columns, for which we sometimes observed worse performance on the same datasets when including more features. We believe this makes our evaluation more challenging and realistic.

\textbf{chemreact, Webb spam, covtype}. For these datasets we directly used the preprocessed versions and code from \cite{huggins2016coresets} (\url{https://bitbucket.org/jhhuggins/lrcoresets/src/master/}). These match standard usage of the datasets, except for Webb spam. This dataset contained 126185 entries instead of 350000 which was reported in their paper. We used the version that was available.

\textbf{KDD cup 99}. We obtained the 10\% subset from \url{http://kdd.ics.uci.edu/databases/kddcup99/kddcup99.html}. Columns 2-4 are categorical which we one-hot encoded. We applied min-max scaling to the range [-1, 1] and reserved 5\% for the test set.

\textbf{bank}. We obtained the dataset from the UCI repository \url{https://archive.ics.uci.edu/ml/datasets/bank+marketing}. We kept the numerical variables age, duration, campaign, previous, emp.var.rate, cons.conf.idx, euribor3m, nr.employed and the categorical variables job, marital, education, default, housing, loan, contact, month, day\_of\_week, poutcome which we one-hot encoded. We applied min-max scaling and reserved 5\% for the test set.

\textbf{census}. We obtained the dataset from the UCI repository \url{https://archive.ics.uci.edu/ml/datasets/census+income}. We identified and kept columns 1, 3, 5, 11, 12, 13 as numeric, and columns 2, 4, 6, 7, 8, 9, 10, 14 as categorical, which we one-hot encoded. We min-max scaled the dataset and reserved 5\% as the test set.

\textbf{SUSY}. This is the most large-scale dataset we used in order to benchmark coresets on a more realistic application scale. We obtained it from \url{https://archive.ics.uci.edu/ml/datasets/SUSY}. All features are numeric so we applied min-max-scaling. As described in their readme, we reserved the final 500k rows as the test set, leaving 4.5 million for training.

\textbf{bitcoin}. This is a dataset of bitcoin ransomware attacks vs benign transactions, obtained from \url{https://archive.ics.uci.edu/ml/datasets/BitcoinHeistRansomwareAddressDataset}. We also selected this dataset because it was one of the larger non-simulated datasets available. We kept the numeric columns length, weight, count, looped, neighbors, income. We also used the year and day columns as categorical: We one-hot encoded the year directly. For day, we binned the days into 12 groups roughly corresponding to months, and one-hot encoded that. We applied min-max scaling and reserved 5\% as the test set.

\subsection{Coreset code sources}
For $k$-means coreset we used their public code at \url{https://bitbucket.org/jhhuggins/lrcoresets/src/master/}. We modified their sampling function to return the entire $(X, y, w)$ coreset so we could fit the subsampled model. 

We found the R package for OSMAC from \url{https://rdrr.io/github/Ossifragus/OSMAC/}. We re-implemented the method in Python and referred to their code to resolve ambiguities.

To the best of our knowledge, no code was available for any other coreset method so we re-implemented them based on their papers.

\subsection{Computational details}

We distributed the coreset experiments over a cluster of 250 CPUs. A single experiment with 50 replications runs relatively quickly, from a few seconds to a few minutes depending on the size of the data. It was important to precompute the full data coefficient MLEs to save on runtime. We used seeds 0-49 for the replications. All models were implemented using numpy and scipy routines. The $k$-means coreset uses Cython which was compiled following the instructions in their code.

\FloatBarrier
\clearpage
\section{$k$-means coreset parameters}

Below we compare the effects of varying $k$ and $R$. We show on two of the original datasets and metric (estimated test NLL) from the $k$-means coreset paper to be comparable to the original and not overfit in our main experiments. As seen, $R=1.0$ is the best, while $k$ does not make much difference.

\begin{figure*}[h]
\centering
\begin{subfigure}{0.99\textwidth}
\centering
\hspace*{\fill} covertype \hfill Webb spam \hspace*{\fill}
\end{subfigure}
\begin{subfigure}{0.99\textwidth}
\begin{turn}{90} 
\phantom{BufferBuffer}Impact of $R$
\end{turn}
    \centering
    \hspace*{\fill}
    \includegraphics[width=0.3\textwidth, trim=0cm 0cm 0cm 0cm]{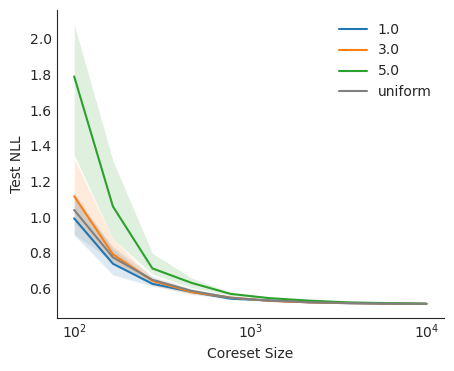}
    \hfill
    \includegraphics[width=0.3\textwidth, trim=0cm 0cm 0cm 0cm]{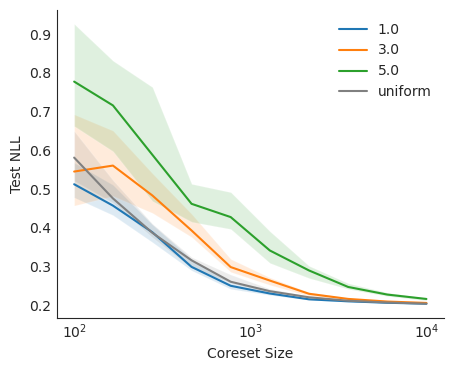}
    \hspace*{\fill}
\end{subfigure}

\begin{subfigure}{0.99\textwidth}
\begin{turn}{90} 
\phantom{BufferBuffer} Impact of $k$
\end{turn}
    \centering
    \hspace*{\fill}
    \includegraphics[width=0.3\textwidth, trim=0cm 0cm 0cm 0cm]{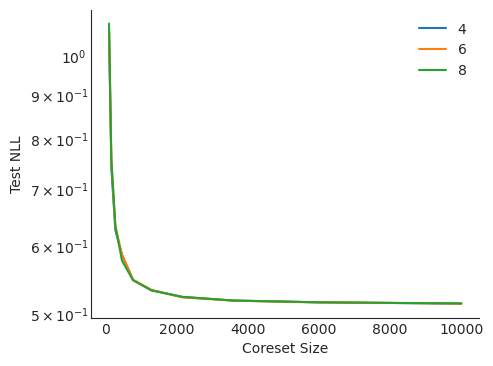}
    \hfill
    \includegraphics[width=0.3\textwidth, trim=0cm 0cm 0cm 0cm]{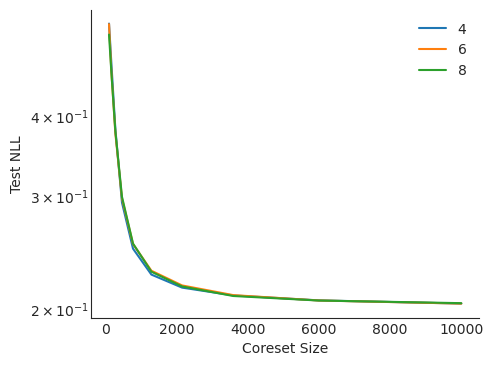}
    \hspace*{\fill}
\end{subfigure}
\label{fig:k_vs_r}
\end{figure*}